\documentclass[letterpaper]{article} 
\usepackage{aaai25}  
\usepackage{times}  
\usepackage{helvet}  
\usepackage{courier}  
\usepackage[hyphens]{url}  
\usepackage{graphicx} 
\usepackage{natbib}  
\usepackage{caption} 
\usepackage{amsmath}
\usepackage{amssymb}

\usepackage{algorithm}
\usepackage{algpseudocode}
\usepackage{mathtools}
\usepackage{lineno}
\usepackage{booktabs}
\usepackage{multirow} 

\usepackage{newfloat}
\usepackage{listings}
\DeclareCaptionStyle{ruled}{labelfont=normalfont,labelsep=colon,strut=off} 

\floatstyle{ruled}
\newfloat{listing}{tb}{lst}{}
\floatname{listing}{Listing}
\title{ProsodyFM: Unsupervised Phrasing and Intonation Control for Intelligible Speech Synthesis}
\author{
    \\
    Xiangheng He\textsuperscript{\rm 1},
    Junjie Chen\textsuperscript{\rm 3},
    Zixing Zhang\textsuperscript{\rm 4}\thanks{Corresponding author},
    Björn W.\ Schuller\textsuperscript{\rm 1,\rm 2}
}

\affiliations{
    \textsuperscript{\rm 1}GLAM -- Group on Language, Audio, \& Music, Imperial College London, UK\\
    \textsuperscript{\rm 2}CHI -- Chair of Health Informatics, MRI, Technical University of Munich, Germany\\
    \textsuperscript{\rm 3}Department of Computer Science, The University of Tokyo, Japan\\
    \textsuperscript{\rm 4}College of Computer Science and Electronic Engineering, Hunan University, China \\
    \{x.he20, bjoern.schuller\}@imperial.ac.uk, junjiechen@g.ecc.u-tokyo.ac.jp, zixingzhang@hnu.edu.cn
}

\usepackage{bibentry}

\begin{document}

\maketitle

\begin{abstract}
Prosody contains rich information beyond the literal meaning of words, which is crucial for the intelligibility of speech. Current models still fall short in phrasing and intonation; they not only miss or misplace breaks when synthesizing long sentences with complex structures but also produce unnatural intonation. We propose ProsodyFM, a prosody-aware text-to-speech synthesis (TTS) model with a flow-matching (FM) backbone that aims to enhance the phrasing and intonation aspects of prosody. ProsodyFM introduces two key components: a Phrase Break Encoder to capture initial phrase break locations, followed by a Duration Predictor for the flexible adjustment of break durations; and a Terminal Intonation Encoder which learns a bank of intonation shape tokens combined with a novel Pitch Processor for more robust modeling of human-perceived intonation change. ProsodyFM is trained with no explicit prosodic labels and yet can uncover a broad spectrum of break durations and intonation patterns. Experimental results demonstrate that ProsodyFM can effectively improve the phrasing and intonation aspects of prosody, thereby enhancing the overall intelligibility compared to four state-of-the-art (SOTA) models. Out-of-distribution experiments show that this prosody improvement can further bring ProsodyFM superior generalizability for unseen complex sentences and speakers. Our case study intuitively illustrates the powerful and fine-grained controllability of ProsodyFM over phrasing and intonation. 
\begin{links}
\link{Code and demo}{https://github.com/XianghengHee/ProsodyFM}
\link{Extended version with Appendix}{https://arxiv.org/abs/2412.11795}
\end{links}

\end{abstract}

%

\section{Introduction}

Prosody, which encompasses various properties of speech such as phrasing, intonation, prominence, and rhythm, can convey rich information beyond the literal meaning of words \cite{xu2019prosody}. It plays a crucial role in the intelligibility of speech. Although recent TTS models have achieved great progress in synthesizing intelligible speech, they still lack in many prosody aspects. In this study, we focus on two prosody aspects in English: phrasing and intonation.

Phrasing refers to grouping words into chunks. An intonational phrase contains a chunk of words with their own intonation pattern. \textbf{Phrase break} in this paper refers to the perceivable acoustic pause at the end of intonational phrases. Phrase break plays an important role in enhancing speech intelligibility \cite{futamata2021phrase}. It implies the phrasal organization in the sentence, allowing listeners to accurately discern the syntactic structure of the sentence and deduce its correct meaning. For example, the sentence ``I saw the man with the telescope." can be interpreted differently depending on whether there is a break after the word ``man". When the sentence is spoken without the break, ``with the telescope" modifies ``the man", suggesting that the man observed by the speaker had a telescope. When the break is introduced after ``man", it implies the speaker used a telescope to see the man. This demonstrates that incorrect phrasing can lead to incorrect interpretation of the sentence, thereby impairing speech intelligibility. However, due to the difficulty in obtaining break labels and the variability of break duration, current TTS systems usually miss or misplace breaks when synthesizing complex sentences.

Intonation, especially terminal intonation, is essential for synthesizing intelligible speech. We refer to the intonation pattern of the last word in an intonational phrase as the \textbf{terminal intonation}. Terminal intonation carries many linguistic and paralinguistic information in English. A rising terminal intonation at the end of a sentence usually signals uncertainty or a request for clarification, while a falling intonation typically indicates certainty or is used to make statements and assertions \cite{liberman1975intonational}. A rising terminal intonation in the middle of a sentence usually indicates the speaker is not finished yet, while a falling tone indicates the end of the thought \cite{bolinger1998intonation}. This information of intonation change can be represented through the change in the pitch contour \cite{cole2022shape}. However, instead of modeling the relative change in the pitch contour, previous TTS systems directly model the absolute pitch value. This design choice hinders their ability to accurately capture natural intonation, as pitch tracking and predicting absolute pitch values is inherently challenging.


To address these issues, we propose ProsodyFM, a novel Prosody-aware TTS model based on a Flow-Matching (FM) backbone that enhances both phrasing and intonation aspects of prosody in an unsupervised manner, resulting in more intelligible synthesized speech. For the break labeling issue, we introduce a Phrase Break Encoder to capture initial break locations, followed by a Duration Predictor to adjust break durations, enabling flexible and accurate modeling of phrase breaks. For the intonation modeling issue, we employ a novel Pitch Processor and learn a bank of intonation shape tokens, which effectively mitigates pitch tracking errors, enables more robust modeling of pitch shapes, and aligns more closely with human perception of intonation changes. ProsodyFM is trained without any prosodic labels and yet can uncover a wide range of break durations and intonation patterns. The main contributions of this paper are as follows:
\begin{itemize}
    \item We propose ProsodyFM, a prosody-aware TTS model with strong generalizability and fine-grained prosody control, capable of synthesizing speech with natural phrasing and intonation, leading to greater intelligibility than existing systems. 

    \item We provide novel and effective solutions for the break labeling issue and the intonation modeling issue.
    
    \item We release our demo, code, and model checkpoints to facilitate further research.
\end{itemize}


\section{Related Works}
\subsection{The Break Labeling Issue}
Breaks in speech can be roughly divided into punctuation-based and respiratory breaks \cite{hwang2023pausespeech}. Unlike punctuation-based breaks which are marked by punctuations, respiratory breaks have no explicit label on the text side. Most of the current TTS systems \cite{mehta2024matcha,li2024styletts} have only considered punctuation-based breaks, resulting in many non-final phrase breaks being overlooked or misplaced \cite{taylor2009text}. Some TTS systems model phrase breaks explicitly. These models \cite{hwang2023pausespeech, Abbas2022ExpressiveVA, yang2023duration} use manually designed thresholds combined with the Montreal Forced Aligner (MFA) \cite{mcauliffe2017montreal} to obtain break labels in an unsupervised manner. The frequency and duration of the phrase break are shaped by both the linguistic phrase structure and a speaker’s speaking style \cite{hwang2023pausespeech}. However, due to the variability of break duration and its dependence on speaker information, the handcraft threshold-based methods can hardly account for speaker-specific variations in break durations. 

ProsodyFM tackles this issue by designing a Fusion Encoder to integrate initial break locations obtained from the Phrase Break Encoder with speaker information, and then adjusting the break durations with a Duration Predictor, enabling flexible modeling of phrase breaks.

\subsection{The Intonation Modeling Issue}
\label{pich_error}
\begin{figure}[t]
  \includegraphics[width=1\linewidth]{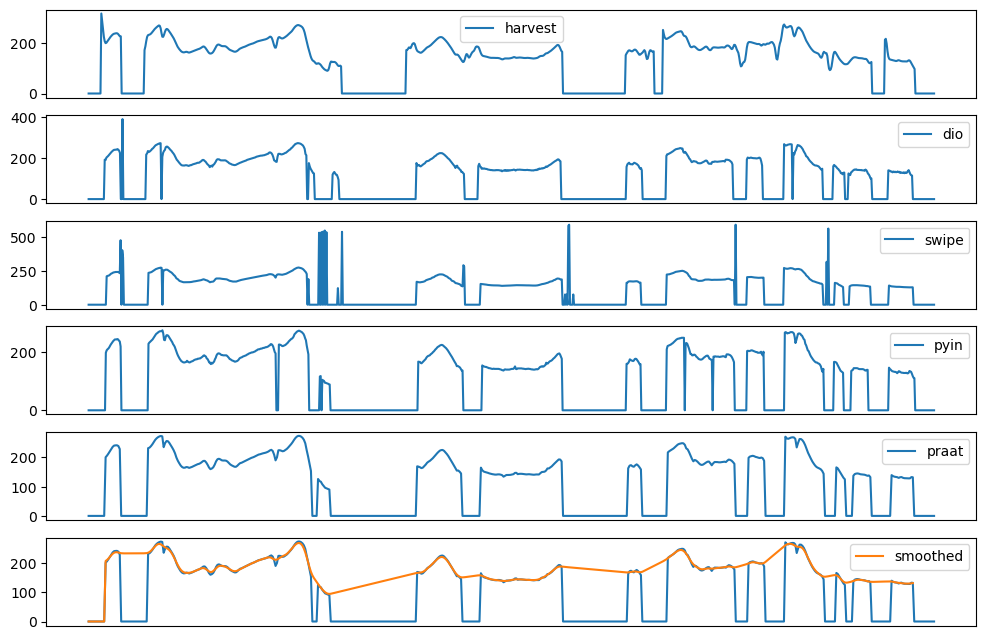} 
  \caption {Pitch contours extracted from 5 pitch tracking methods (blue) and our pitch smoothing method (orange).} 
  \label{pitch_tracking}
  \vspace{-0.5cm}
\end{figure}

Annotating intonation pattern labels is a high-cost task and often yields unreliable results \cite{lee2019robust} due to the complexity of current annotation systems \cite{silverman1992tobi}. Almost all the existing intonation-aware TTS systems \cite{ren2020fastspeech,min2021meta,huang2022generspeech,li2024styletts} directly model the absolute pitch values obtained from some pitch tracking methods. However, pitch tracking is inherently challenging, and existing methods frequently yield errors like pitch doubling/halving and incorrect unvoiced/voiced flags \cite{hirst2021measuring}, leading to unreliable results. Figure \ref{pitch_tracking} illustrates the pitch tracking results across 5 different methods. From top to bottom are Harvest \cite{morise2017harvest}, DIO \cite{morise2009fast}, SWIPE \cite{camacho2008sawtooth}, pYIN \cite{mauch2014pyin}, and Praat \cite{boersma2001praat}. 
We can clearly observe frequent prediction errors in pitch values and unvoiced/voiced flags, as well as inconsistencies across these five methods. Some recent findings from human perceptual studies offer a potential basis for this issue; the authors in 
\cite{chodroff2019testing,cole2022shape} have shown that \textbf{compared to the detailed pitch values, the shape of the pitch contour is more important for human perception of intonation change}.

ProsodyFM introduces a novel Pitch Processor that interpolates, smooths, and perturbs raw pitch values to highlight their shape, and subsequently learns a set of intonation shape tokens to model perceptually aligned intonation change instead of directly modeling absolute pitch values. The orange line in Figure \ref{pitch_tracking} shows an example after our smoothing process. Our method alleviates pitch tracking errors, enables more robust modeling of pitch shapes, and aligns more closely with human perception of intonation change.



\section{Method}
\begin{figure*}[t]
  \includegraphics[width=1\linewidth]{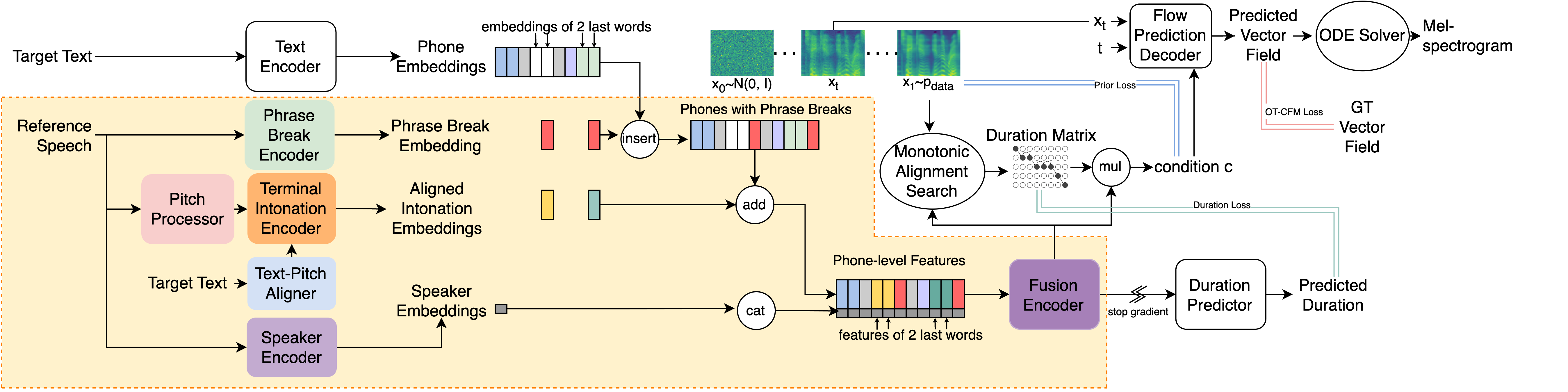} 
  \caption {The model architecture of the proposed ProsodyFM during training. The components outlined by the yellow shaded area are unique to ProsodyFM and differ from those in MatchaTTS.}
  \label{model}
\end{figure*}

\begin{figure*}[t]
  \vspace{-0.5cm}
  \includegraphics[width=1\linewidth]{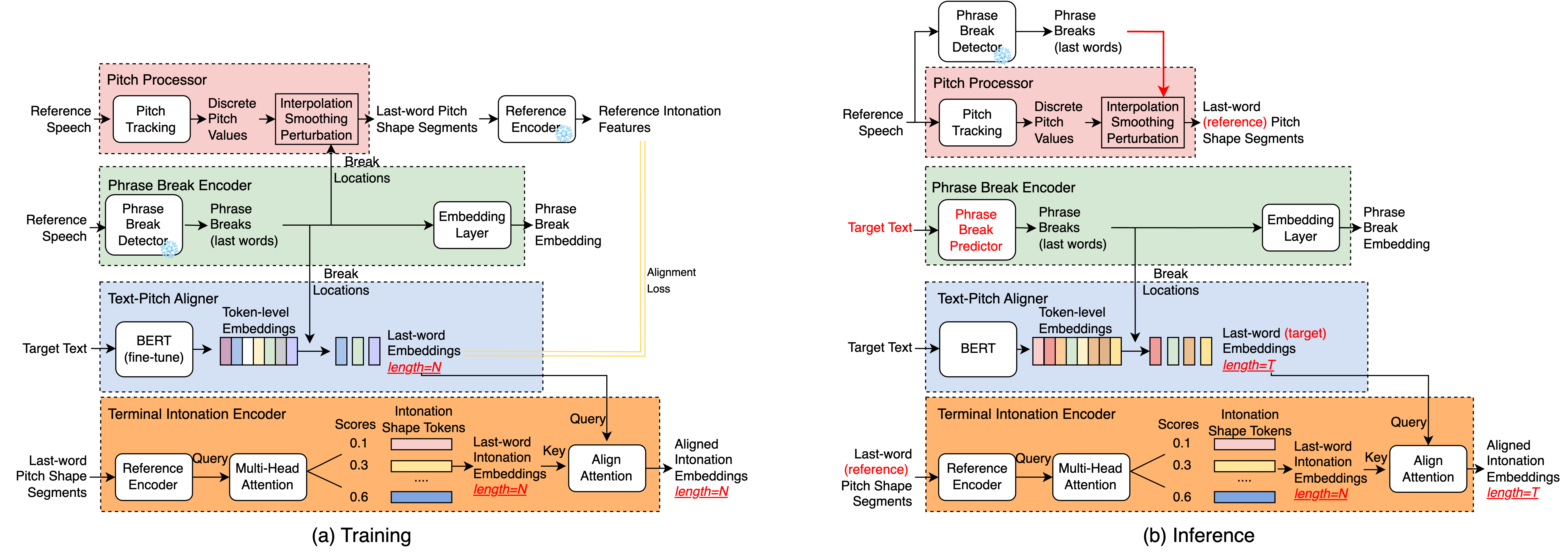} 
  \caption {The key components of the proposed ProsodyFM in the training (a) and inference (b) phrases. The red markings highlight the differences. The snowflake mark means the module is frozen during training.}
  \vspace{-0.5cm}
  \label{model_components}
\end{figure*}

ProsodyFM is designed to extract phrasing and terminal intonation patterns from reference speech and adjust these patterns to match the target text. Following the MatchaTTS \cite{mehta2024matcha} backbone, ProsodyFM is trained using the Optimal-Transport Conditional Flow Matching (OT-CFM) \cite{lipman2022flow}. The formulation and training algorithm of ProsodyFM can be found in Appendix A of the extended version. ProsodyFM predicts Mel-spectrograms from raw text, which are then converted to waveforms using the HifiGAN vocoder \cite{kong2020hifi}.

The given target text aligns with the reference speech during training but may differ during inference. During training, the reference speech serves as the ground truth and the target text matches its transcript, while during inference, the target text may not match the transcript of the reference speech. 

Figure~\ref{model} illustrates the overall structure of ProsodyFM, highlighting the proposed components within the yellow-shaded area. Details of four key components are presented in Figure~\ref{model_components}: (1) the Pitch Processor extracts robust pitch shape segments; (2) the Phrase Break Encoder predicts initial phrase break locations, which are then combined with speaker information and refined for duration by the Duration Predictor; (3) the Text-Pitch Aligner estimates intonation patterns from the target text to guide the selection of reference intonation patterns; and (4) the Terminal Intonation Encoder models terminal intonation patterns that are properly aligned with the target text.

\subsection{Pitch Processor}
The Pitch Processor (pink box in Figure \ref{model_components}) extracts robust pitch shape segments of the last words through three operations: interpolation, smoothing, and perturbation. First, it interpolates and smooths the discrete, unreliable raw pitch values from pitch tracking into continuous contours. Then, to emphasize pitch shape over absolute values, it subtracts a random offset (uniformly sampled from \([f_{min}, f_{max}]\)) from each contour point, preserving the shape patterns while perturbing its specific value information.


\subsection{Phrase Break Encoder}
The Phrase Break Encoder (green box in Figure \ref{model_components}) predicts where phrase breaks occur, thus allowing it to locate the last word of each intonational phrase. These last-word locations guide the Pitch Processor and the Text-Pitch Aligner in selecting the corresponding pitch shape segments and word embeddings.

During training, the Phrase Break Encoder uses a pre-trained, frozen Phrase Break Detector to identify phrase breaks from reference speech. During inference, when no aligned reference speech is available, the Phrase Break Encoder relies on a Phrase Break Predictor fine-tuned from T5 \cite{ni-etal-2022-sentence} to infer breaks directly from plain target text. The performance of this Phrase Break Predictor is reported in Appendix B of the extended version.


\subsection{Text-Pitch Aligner}
The Text-Pitch Aligner (blue box in Figure \ref{model_components}) predicts intonation patterns from the target text, even without matched speech during inference. We fine-tune BERT by minimizing the L2 loss between BERT-derived word embeddings and the reference intonation features extracted by the Reference Encoder. The Reference Encoder is identical to the one in the Terminal Intonation Encoder, but detached to prevent gradient flow. The predicted BERT embeddings then guide the selection of suitable reference intonation patterns in the Terminal Intonation Encoder.

\subsection{Terminal Intonation Encoder}

The Terminal Intonation Encoder (orange box in Figure \ref{model_components}) extracts the terminal intonation patterns that are aligned with the target text. The Reference Encoder compresses the pitch shape segments of the last word in the reference speech into a fixed-length intonation feature, used as the query for the Multi-head Attention module. This attention module learns a similarity measure between the reference intonation features and a bank of \textbf{intonation shape tokens}. These tokens serve as a learnable codebook designed to capture and represent various intonation patterns. Trained with OT-CFM loss alone, these tokens require no annotated intonation labels. The Multi-head Attention module generates weights for these tokens, and their weighted sum forms the last-word intonation embedding of the reference speech.

However, during inference, the reference speech may not be aligned with the target text, resulting in a different number of last words in the reference speech compared to the target text. We use scaled dot-product attention (Align Attention module) to select the terminal intonation patterns from the reference speech that best suit the target text. Specifically, we treat the last-word intonation embeddings (of the reference speech) as the key (and value) and the last-word embeddings (of the target text) as the query. This alignment enables ProsodyFM to autonomously choose the terminal intonation pattern based on both the reference speech and the target text during the inference phase.

\subsection{Mel-spectrogram Generation}
During inference, the Fusion Encoder combines the phrase break and aligned intonation embeddings with speaker and phone embeddings to produce phone-level prior statistics. The Duration Predictor (instead of the MAS during training) then determines the optimal durations of each phone and phrase break to obtain the frame-level condition \(c\). Given \(c\), a sampled time \(t\), and \(x_t\), the Flow Prediction Decoder predicts the target vector field. Finally, the ODE solver uses this predicted vector field to generate the Mel-spectrogram.


\section{Experimental Details}
\label{details}
\subsubsection{Model Configurations} For a fair comparison, we utilize the same model architecture and hyperparameters as MatchaTTS \cite{mehta2024matcha} except for the following modules. For our Terminal Intonation Encoder, we employ the attention module in \cite{wang2018style} with 4 attention heads and 6 64-D tokens. We replace the complex reference encoder in \cite{wang2018style} with a single-layer LSTM with 128-D hidden size to speed up training. For the Phrase Break Detector, we use the released checkpoint of PSST \cite{roll-etal-2023-psst} without fine-tuning. For the Phrase Break Predictor, we fine-tune T5 \cite{ni-etal-2022-sentence} independent from ProsodyFM using LoRA \cite{hu2021lora} with 16 ranks and consider the phrase breaks obtained from the PSST as the ground truth labels when fine-tuning. For the Text-Pitch Aligner, we initialize BERT\footnote{https://huggingface.co/google-bert/bert-base-uncased} with pre-trained weights, using its original tokenizer to process input text and obtain 768-D token-level embeddings. We then select the tokens corresponding to each last word, average their embeddings, and pass it through a fully connected layer to produce a final 192-D embedding for each last word. 
For the Pitch Processor, we use Praat \cite{boersma2001praat} to extract discrete pitch values and use a customized Praat script modified from \cite{cangemi2015} to interpolate and smooth them into a continuous pitch contour. For the Speaker Encoder, we extract the same external speaker embedding as in \cite{casanova2022yourtts} for each speech sample and add two fully connected layers to transform the 512-D d-vector to the final 64-D speaker embeddings.

\subsubsection{Datasets} We perform the experiments in Table \ref{performance_objective}, Table \ref{performance_subjective}, Table \ref{ablation}, and Figure \ref{case} on the LibriTTS corpus \cite{zen2019libritts}. We randomly split (speakers-independent) the audio samples in the train-clean-100, dev-clean, and test-clean sections of LibriTTS into 40421, 839, and 839 samples for our training, validation, and testing sets, respectively. The whole dataset has in total 71 hours of audio signals and 326 speakers. For the experiments in Table \ref{generalizability}, we train the models on the VCTK corpus \cite{yamagishi2019vctk} with the same training set as in \cite{kim2021conditional} and test the models on our LibriTTS testing set. We resample all audio to 22050\,Hz and extract Mel-spectrograms with a 1024 FFT size, 256 hop size, 1024 window length, and 80 frequency bins.

\subsubsection{Training Settings} ProsodyFM and its ablated variants in Table \ref{ablation} are trained for 350 epochs on an NVIDIA A100 GPU with 80GB VRAM with batch size 64 and learning rate 1e-4. 

\subsubsection{Objective Evaluation Metrics} 
We conduct objective evaluations with the log-scale F0 Root Mean Squared Error ($RMSE_{f0}$), the F1 score of the break classification ($F1_{break}$), and the Word Error Rate ($WER$).

1) $RMSE_{f0}$ measures the pitch error. Following \cite{birkholz2020accounting}, we use it to evaluate the \textbf{intonation} aspect of prosody. We leverage dynamic time warping (DTW) to extract the pitch values and measure the portion where both the ground truth speech $y_i$ and synthesized speech $y'_i$ are voiced.
\begin{equation}
\begin{split}
\text{RMSE}_{f_0} = \sqrt{\frac{1}{T} \sum_{i=1}^{T} \left( \log \left(\frac{y_i}{y'_i}\right)\right)^2}
\end{split}
\end{equation} 
where \(T\) refers to the number of voiced frames.

2) $F1_{break}$ evaluates the \textbf{phrasing} aspect of prosody. We use the PSST \cite{roll-etal-2023-psst} to obtain phrase breaks from the ground truth speech as labels, then apply PSST to the synthesized speech to detect phrase breaks and calculate the F1 score.

3) $WER$ correlates well with the \textbf{intelligibility} of synthesised speech \cite{taylor2021confidence, mehta2024matcha}. We use Whisper-small\footnote{https://huggingface.co/openai/whisper-small} \cite{radford2023robust} to obtain transcripts of both the ground truth and synthesized speech and compute the $WER$.

\subsubsection{Subjective Evaluation Metrics} 
We conduct a crowd-sourced Mean Opinion Score (MOS) human listening test to assess four aspects of synthesized speech, including the phrase break similarity ($MOS_{break}$), the terminal intonation similarity ($MOS_{intonation}$) between the synthesized and a reference speech, the intelligibility ($MOS_{intelligibility}$) and the quality ($MOS$) of the synthesized speech. Each MOS is assessed using a 5-point scale with 95\% confidence intervals, where score 1 indicates dissimilarity, unintelligibility, or poor quality whereas score 5 signifies full similarity, intelligibility, or excellent quality. We randomly select 15 utterances (3 groups of 5) with different lengths in the testing set, and each sample is rated by 21 testers. Our testers are PhD students from four universities specializing in Computer Audition, with native languages including English, German, Chinese, and Turkish. They are all fluent in English. 

To account for the possibility of non-expert testers, we provided detailed explanations of the four MOS metrics with clear definitions and examples before starting the test. The instruction page is in Appendix D of the extended version.

Considering text consistency between the reference audio and the target text to be synthesized, we perform both parallel and non-parallel MOS tests. Given the subjective nature of prosody evaluation, where individuals from different linguistic backgrounds can have varying perceptions of what constitutes `appropriate' phrasing and intonation for a target text \cite{grover1987intonation}, we adopt specific assumptions for our parallel and non-parallel subjective evaluations:

1) For the parallel subjective evaluation, we assume that the phrasing and intonation derived from the reference speech represent the `appropriate' prosody. We provide labels for breaks and intonation based on the reference speech and ask testers to assess the \textbf{similarity} of the phrasing and intonation to these labels, rather than their appropriateness.

2) For the non-parallel subjective evaluation, a reference speech is still needed for similarity assessment. To maintain the relevance of prosody while accommodating different text content, we modify the words in the reference speech (the same 15 samples) transcript, ensuring that the sentence semantics and structure remain as close as possible to the original. We then transfer the break and intonation labels from the reference speech to the new target text. Testers are again asked to assess the \textbf{similarity} of the phrasing and intonation to these labels, rather than their appropriateness. This evaluation rests on the assumption that two sentences with similar semantics and structure should share same phrasing and intonation labels.

The target text and transcripts of reference speech with break and intonation labels under parallel and non-parallel settings can be found in Appendix E of the extended version.

\subsubsection{Comparative Models} 
To evaluate the performance of our model, we compare ProsodyFM with four SOTA models\footnote{The checkpoints used are from their official implementations.}: (1) StyleSpeech \cite{min2021meta}: the expressive multi-speaker TTS model built on FastSpeech2 \cite{ren2020fastspeech}; (2) GenerSpeech \cite{huang2022generspeech}: the TTS model towards high-fidelity style transfer, also extended from FastSpeech2 \cite{ren2020fastspeech}; (3) StyleTTS2 \cite{li2024styletts}: the expressive TTS model with human-level speech quality, improved from StyleTTS \cite{li2022styletts}; (4) MatchaTTS \cite{mehta2024matcha}: the fast and high-quality TTS model based on conditional flow matching. 

To verify the effectiveness of our proposed modules, we compare ProsodyFM against three ablated variants: (5) w/o\_intonation: remove the Terminal Intonation Encoder from ProsodyFM; (6) w/o\_break: remove the Phrase Break Encoder from ProsodyFM; (7) w/o\_into\_break: remove both the Terminal Intonation Encoder and the Phrase Break Encoder from ProsodyFM.

To provide a reference upper bound, we also include: (8) GT(vocoder): we extract the Mel-spectrogram from the ground truth audio and then reconstruct it using HiFiGAN.

\section{Results}
\begin{table}
\begin{tabular}{c|ccc}
\toprule[1.5pt]
Models & $RMSE_{f0}\downarrow$ & $WER\downarrow$ & $F1_{break}\uparrow$  \\
\midrule[1pt]
GT(vocoder)             & 0.2264                   & 2.05\%               & 77.02                    \\
\midrule[1pt]
StyleSpeech             & 0.4477                   & 4.30\%               & 60.49                    \\
GenerSpeech             & 0.4556                   & 7.29\%               & 60.30                    \\
StyleTTS2               & 0.3120                   & 3.25\%               & 58.44                    \\
MatchaTTS               & 0.3376                   & 3.56\%               & 60.08                    \\
\midrule[1pt]
ProsodyFM             & \textbf{0.3068}          & \textbf{3.22\%}      & \textbf{62.76}   
\\
\bottomrule[1.5pt]
\end{tabular}
\caption{Objective results on the LibriTTS testing set. }
\vspace{-0.5cm}
\label{performance_objective}
\end{table}

\begin{table*}[t]
\begin{tabular}{c|cc|cc|cc|cc}

\toprule[1.5pt]
\multirow{2}{*}{Models} & \multicolumn{2}{c|}{$MOS$}                 & \multicolumn{2}{c|}{$MOS_{intelligibility}$}               & \multicolumn{2}{c|}{$MOS_{intonation}$}            & \multicolumn{2}{c}{$MOS_{break}$}               \\
                        & Parallel           & Non-para       & Parallel           & Non-para       & Parallel           & Non-para       & Parallel           & Non-para       \\
\midrule[1pt]
GT(vocoder)             & 4.77±0.05          & -                  & 4.77±0.05          & -                  & 4.90±0.04          & -                  & 4.91±0.04          & -                  \\
\midrule[1pt]
StyleSpeech             & 3.43±0.10          & 3.39±0.10          & 3.72±0.11          & 3.72±0.11          & 3.58±0.10          & 3.53±0.10          & 3.52±0.12          & 3.74±0.11          \\
GenerSpeech             & 2.67±0.10          & -                  & 3.03±0.12          & -                  & 3.14±0.12          & -                  & 2.85±0.13          & -                  \\
StyleTTS2               & 4.09±0.10  & \textbf{4.26±0.10} & 4.23±0.10          & 4.36±0.09          & 3.66±0.10          & 3.78±0.11          & 3.82±0.12          & 4.08±0.11          \\
MatchaTTS               & 3.61±0.10          & 3.90±0.09          & 4.03±0.10          & 4.10±0.10          & 3.48±0.10          & 3.73±0.10          & 3.88±0.10          & 4.08±0.11          \\
\midrule[1pt]
ProsodyFM               & \textbf{4.22±0.08}          & 4.07±0.09          & \textbf{4.47±0.07} & \textbf{4.40±0.08} & \textbf{4.36±0.08} & \textbf{4.24±0.09} & \textbf{4.42±0.08} & \textbf{4.34±0.09} \\
\bottomrule[1.5pt]
\end{tabular}
\caption{MOS results with 95\% confidence intervals on the LibriTTS testing set. ``Parallel'' and  ``Non-para'' indicate that the transcript of the reference audio is the same with or different from the target text, respectively.}
\label{performance_subjective}
\vspace{-0.5cm}
\end{table*}

\subsection{Model Performance}
We conduct both objective and subjective evaluations to assess ProsodyFM and four SOTA models in terms of phrasing (break), intonation, and overall intelligibility. Table \ref{performance_objective} and Table \ref{performance_subjective} show the results.

\subsubsection{Objective Results} As shown in Table \ref{performance_objective}, we observe that ProsodyFM outperforms the other four SOTA models across all three objective evaluation metrics. 
Additionally, the results for phrasing ($F1_{break}$) and intonation ($RMSE_{f0}$) show a positive correlation with overall intelligibility ($WER$). These results indicate that ProsodyFM exhibits superior performance in phrasing and intonation, which further contributes to its enhanced intelligibility.

\subsubsection{Subjective Results}
As shown in Table \ref{performance_subjective}, compared to the other four SOTA models, ProsodyFM obtains significantly better scores in terms of $MOS_{intonation}$ and $MOS_{break}$ under both parallel and non-parallel settings; it also achieves significantly better $MOS_{intelligibility}$ under the parallel setting. In the non-parallel setting, ProsodyFM matches the $MOS_{intelligibility}$ of StyleTTS2 and surpasses the remaining three models significantly. In the non-parallel setting, ProsodyFM shows lower speech quality ($MOS$) than StyleTTS2, likely due to the smaller dataset used in ProsodyFM (71 hours) compared to StyleTTS2 (245 hours). Consistent with the objective evaluation results, we can still observe a positive correlation between $MOS_{intonation}$, $MOS_{break}$ and $MOS_{intelligibility}$, further substantiating that ProsodyFM can effectively improve the phrasing and intonation aspects of prosody, thereby enhancing the speech intelligibility.

\subsection{Model Generalizability}

\begin{table}[]
\centering
\begin{tabular}{c|ccc}
\toprule[1.5pt]
Models      & $RMSE_{f0}\downarrow$          & $WER\downarrow$           & $F1_{break}\uparrow$  \\
\midrule[1pt]
GT(vocoder) & 0.2264          & 2.05\%          & 77.02\% \\
\midrule[1pt]
MatchaTTS   & 0.4727          & 6.78\%          & 55.28\%\\
ProsodyFM   & \textbf{0.4080} & \textbf{4.97\%} & \textbf{59.95\%} \\
\bottomrule[1.5pt]
\end{tabular}
\caption{Objective evaluation results on the out-of-distribution (unseen long and complex sentences, unseen speakers) testing data. All the models are trained on the VCTK (short sentences) training set and tested on the LibriTTS (long and complex sentences) testing set.}
\label{generalizability}
\vspace{-0.5cm}
\end{table}

To evaluate the impact of enhanced phrasing and intonation on the generalizability of models, we conduct out-of-distribution experiments on unseen complex sentences.
Both MatchaTTS and ProsodyFM are trained on the same VCTK training set (short sentences) and tested on the LibriTTS testing set (long sentences). The speakers in the testing set are unseen during training. Table \ref{generalizability} presents the results.

We observe that although both MatchaTTS and ProsodyFM experience performance declines on the out-of-distribution testing set, the decrease in ProsodyFM's performance is considerably smaller than that of MatchaTTS. Notably, for the $F1_{break}$ metric, ProsodyFM in the out-of-distribution setting achieves matching performance with the four SOTA models in the in-distribution setting (as shown in Table \ref{performance_objective}). This indicates that enhanced phrasing and intonation can bring strong generalizability to the ProsodyFM model.

\subsection{Ablation Study}

\begin{table}[]
\centering
\begin{tabular}{l|ccc}
\toprule[1.5pt]
Models                   & \multicolumn{1}{c}{$RMSE_{f0}\downarrow$} & \multicolumn{1}{c}{$WER\downarrow$} & \multicolumn{1}{c}{$F1_{break}\uparrow$} \\
\midrule[1pt]
GT(vocoder)              & 0.2244                      & 2.01\%                    & 76.04\%                    \\
\midrule[1pt]
ProsodyFM                & \textbf{0.3047}             & \textbf{2.86\%}           & \textbf{62.51\%}           \\
w/o\_intonation           & 0.3373                      & 3.13\%                    & 61.25\%                    \\
w/o\_break                & 0.3355                      & 3.10\%                    & 61.06\%                    \\
w/o\_into\_break & 0.3391                      & 3.33\%                    & 60.25\%   
\\
\bottomrule[1.5pt]
\end{tabular}
\caption{Ablation study on the LibriTTS validation set.}
\vspace{-0.3cm}
\label{ablation}
\end{table}

To verify the necessity and effectiveness of our proposed Phrase Break Encoder and Terminal Intonation Encoder, we compare ProsodyFM against its three ablated variants. Table \ref{ablation} shows the results on our LibriTTS validation set.

We observe an improved $F1_{break}$ score in both w/o\_intonation and w/o\_break compared to w/o\_into\_break, which suggests that both the break encoder and intonation encoder inject partial phrase break information. Those break information may be complementary, leading to a further boost in $F1_{break}$ when we combine the two encoders in ProsodyFM.  The $RMSE_{f0}$ for three ablated variants exhibit no substantial differences, which likely stems from the $RMSE_{f0}$ metric being calculated only for voiced segments. When both intonation and break information are present, ProsodyFM achieves considerably better $RMSE_{f0}$. These observations suggest that both the Phrase Break Encoder and Terminal Intonation Encoder are essential for synthesizing highly intelligible speech.


\begin{figure*}[ht]
\centering
\begin{minipage}{.32\textwidth}
    \centering
    \includegraphics[width=\linewidth]{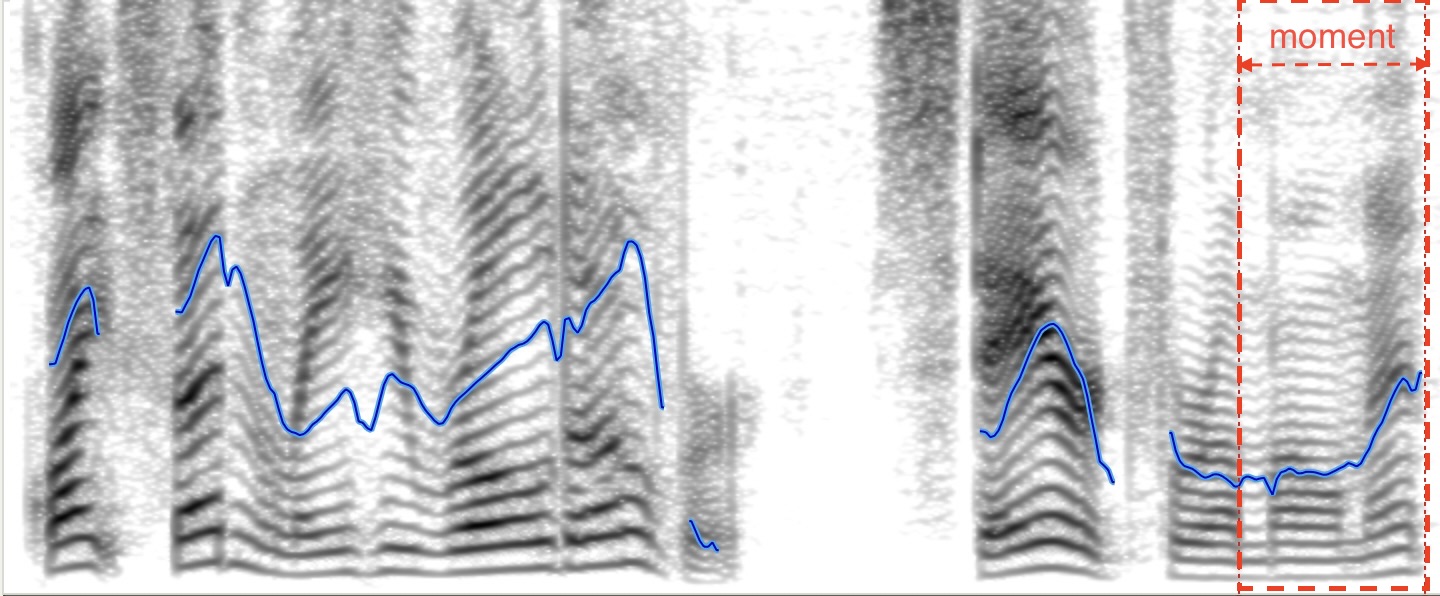}
    \caption*{(a) GT(vocoder)}
\end{minipage}\hfill
\begin{minipage}{.32\textwidth}
    \centering
    \includegraphics[width=\linewidth]{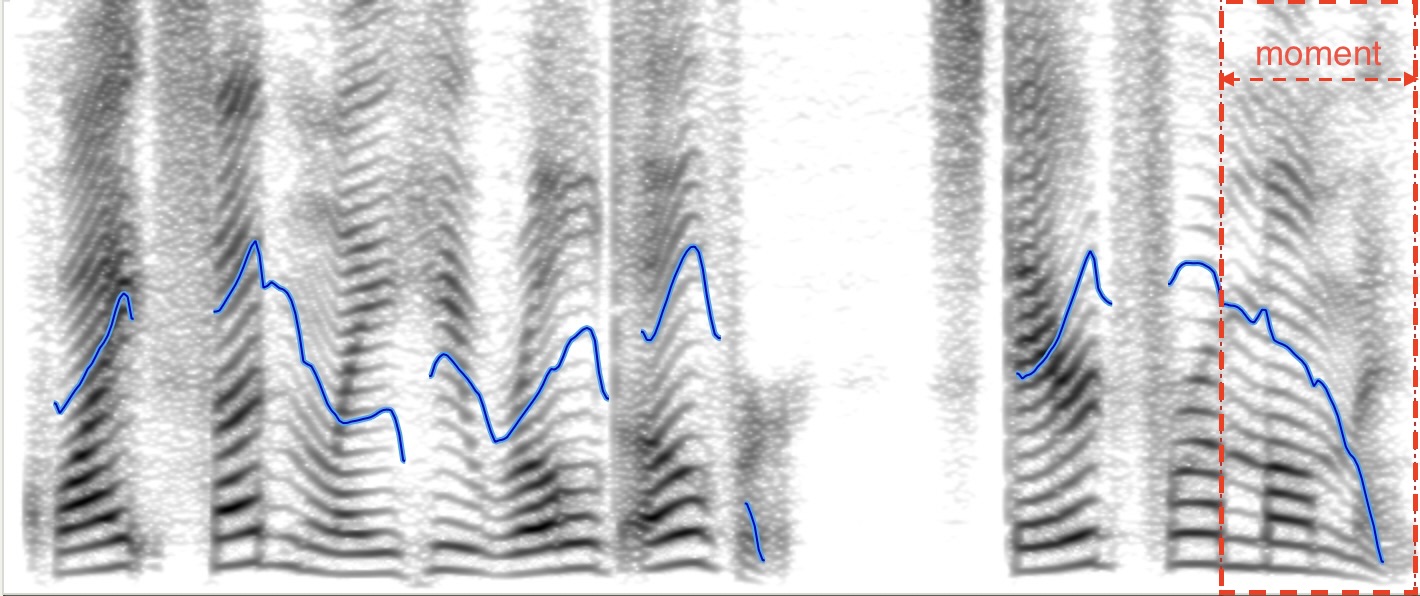}
    \caption*{(d) Falling k=-4}
\end{minipage}\hfill
\begin{minipage}{.32\textwidth}
    \centering
    \includegraphics[width=\linewidth]{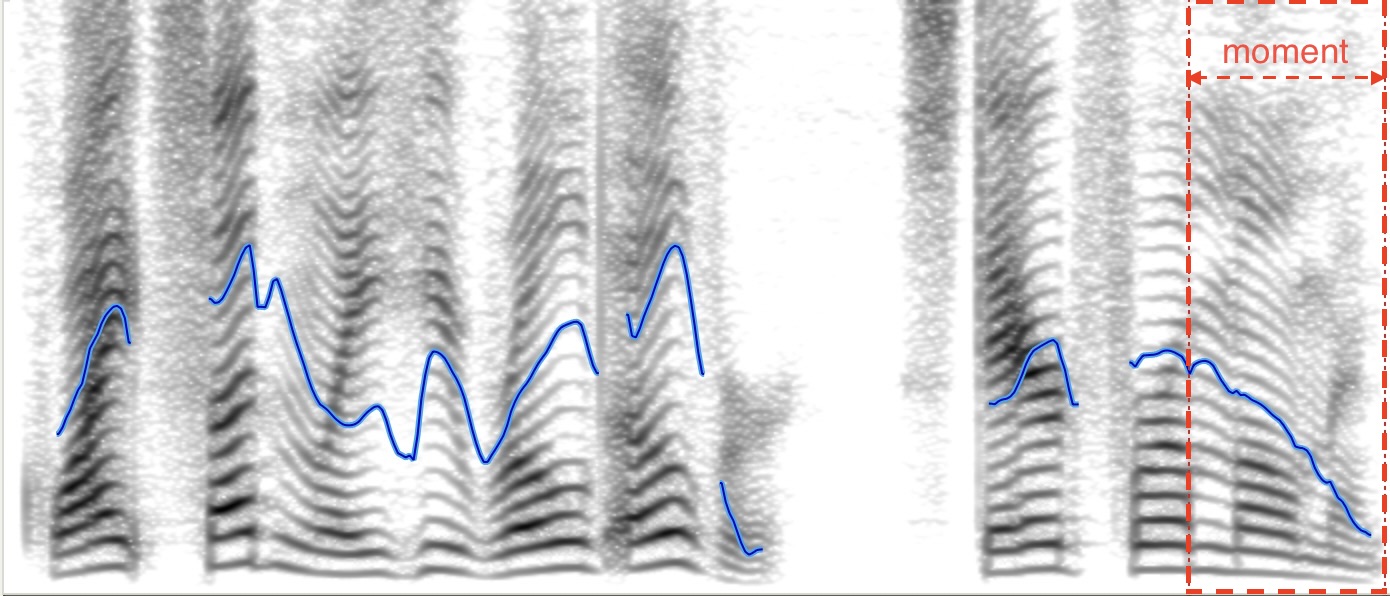}
    \caption*{(g) Falling k=-2}
\end{minipage}

\begin{minipage}{.32\textwidth}
    \centering
    \includegraphics[width=\linewidth]{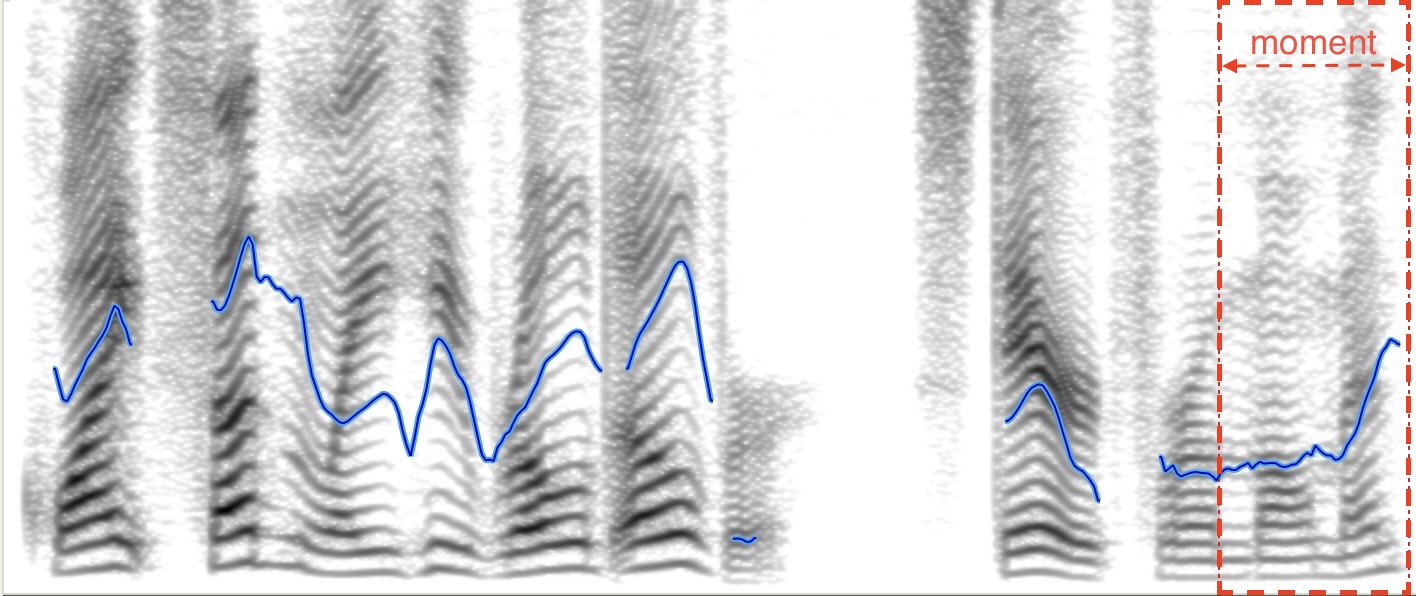}
    \caption*{(b) Prosodyfm (w\_o control)}
\end{minipage}\hfill
\begin{minipage}{.32\textwidth}
    \centering
    \includegraphics[width=\linewidth]{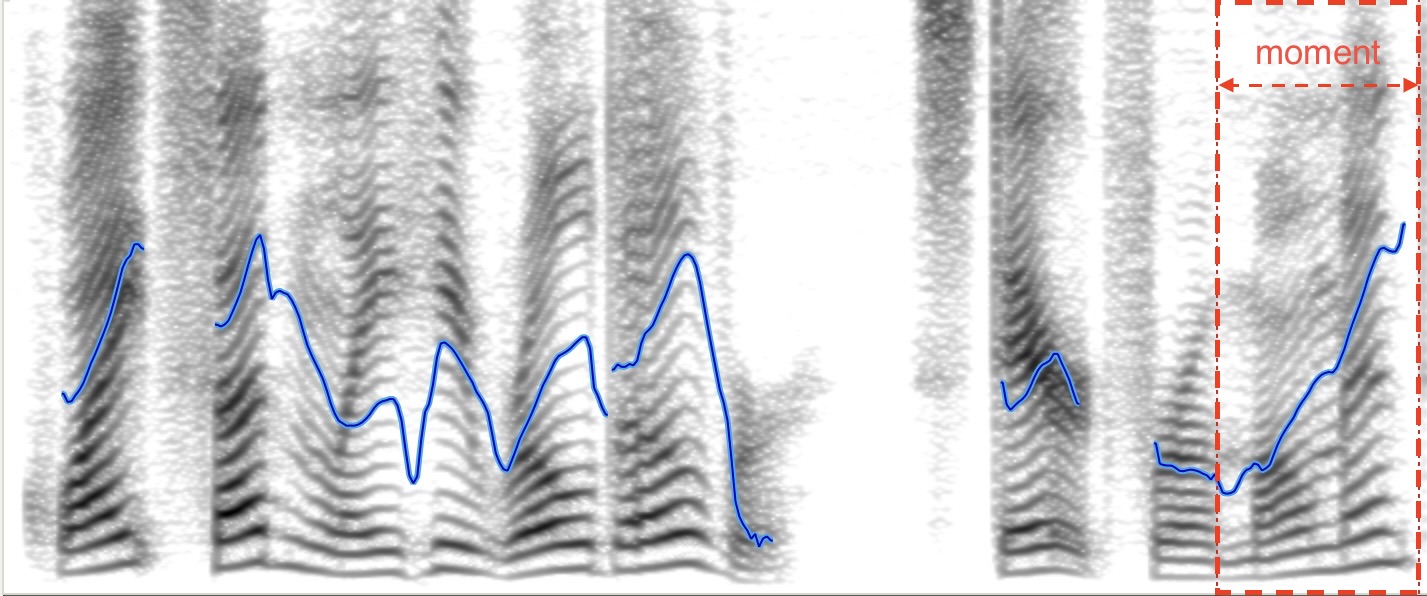}
    \caption*{(e) Rising k=+4}
\end{minipage}\hfill
\begin{minipage}{.32\textwidth}
    \centering
    \includegraphics[width=\linewidth]{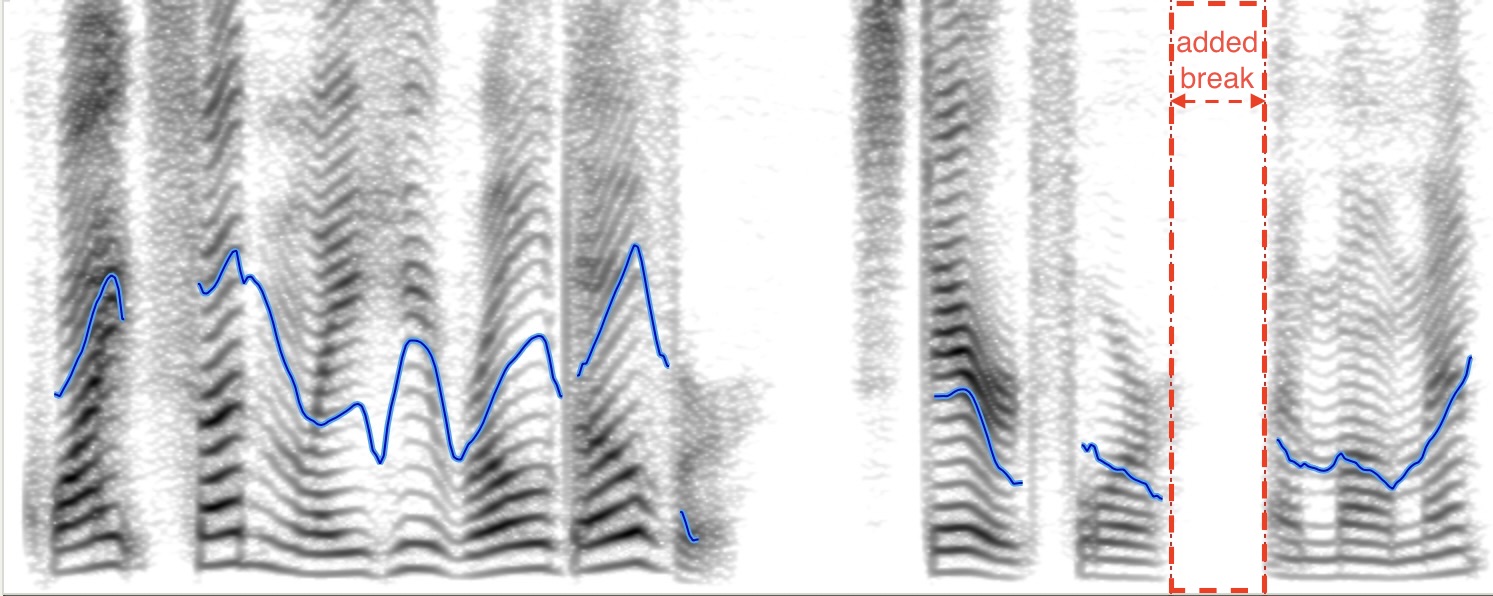}
    \caption*{(h) Add a break}
\end{minipage}

\begin{minipage}{.32\textwidth}
    \centering
    \includegraphics[width=\linewidth]{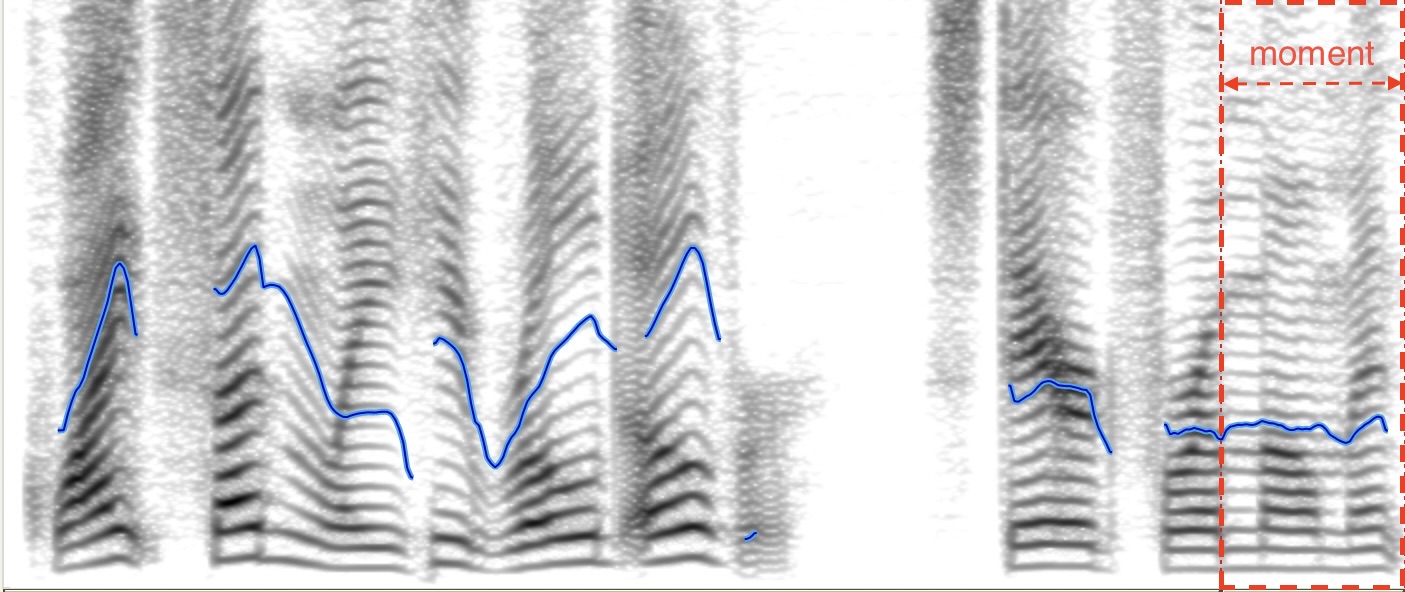}
    \caption*{(c) Level k=0}
\end{minipage}\hfill
\begin{minipage}{.32\textwidth}
    \centering
    \includegraphics[width=\linewidth]{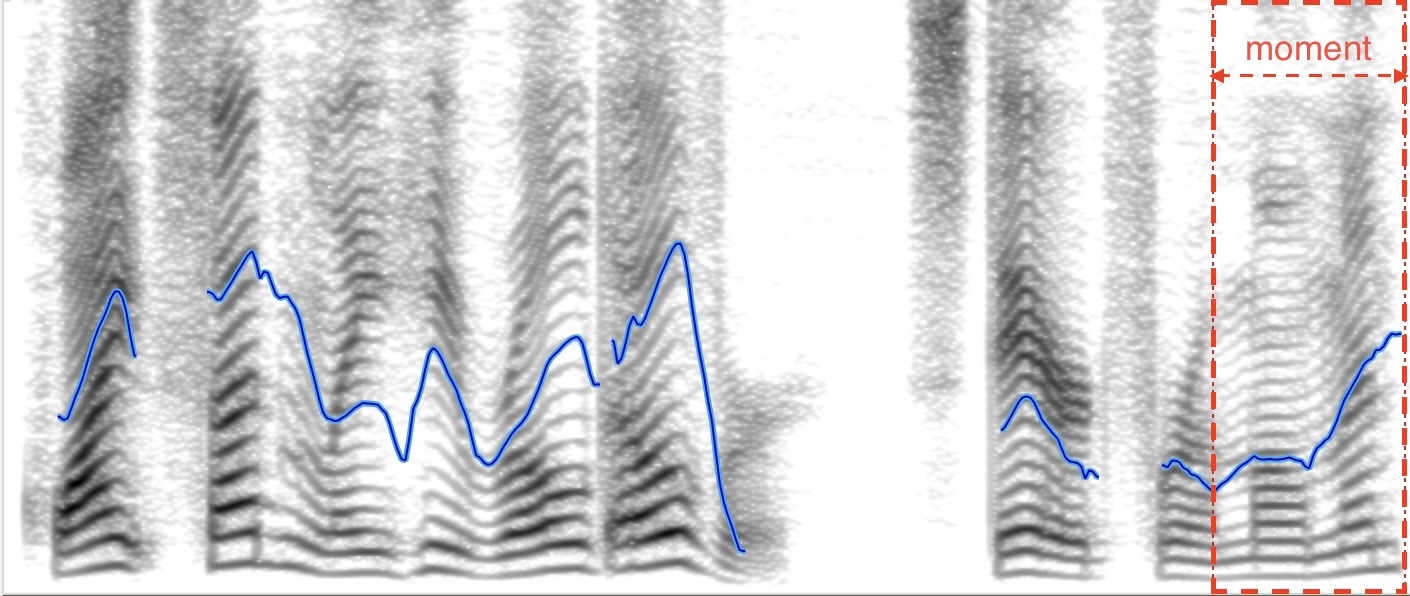}
    \caption*{(f) Rising k=+2}
\end{minipage}\hfill
\begin{minipage}{.32\textwidth}
    \centering
    \includegraphics[width=\linewidth]{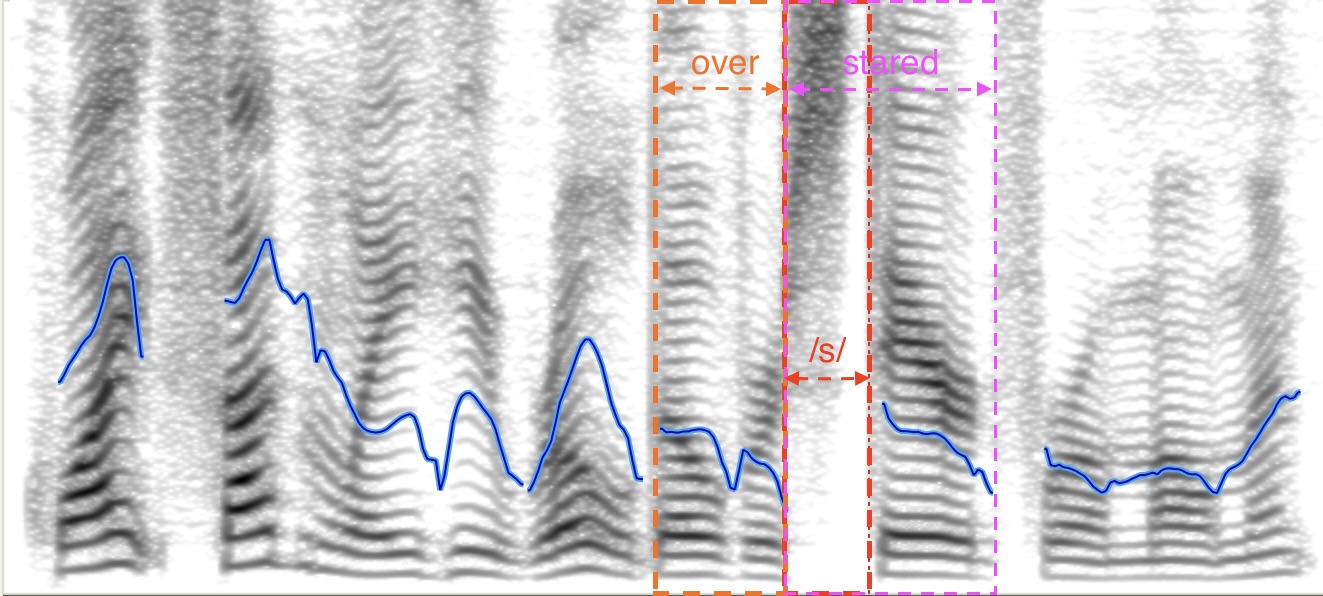}
    \caption*{(i) Remove a break}
\end{minipage}

\caption{Part of spectrograms and pitch contours (blue line) of a reference speech  (1363\_139304\_000009\_000005.wav) and speech synthesized by ProsodyFM with controlled intonation and phrasing.}
\vspace{-0.5cm}
\label{case}
\end{figure*}

\subsection{Case Study: Prosody Controllability}
We present a case study to visually demonstrate the ability of ProsodyFM to control prosody, specifically in terms of intonation and phrasing, which are the focus of this paper. Figure \ref{case} displays the spectrograms of audio samples synthesized with controlled intonation and phrasing (can be found on our demo page). The reference speech is from our LibriTTS testing set. Due to the space limit, we here show part of the spectrograms in Figure \ref{case}. The corresponding transcript with the original break and intonation labels is ``Quite suddenly he rolled \underline{over} (falling tone) stared for a \underline{moment} (rising tone)". We also provide more cases in Appendix C of the extended version.


\subsubsection{Intonation Control} For the controllability of the terminal intonation, we manually modify the reference pitch shape segment of the last word in an intonational phrase and synthesize the corresponding speech using ProsodyFM (we modify the input “Last-word (reference) Pitch Shape Segments” of the Terminal Intonation Encoder in Figure \ref{model_components} (b)). We apply linear adjustments to the pitch values to create rising, falling, and level tones, controlling the magnitude of these adjustments through the slope: a slope of \(k = +4\) represents a rapid rising tone, \(k = +2\) represents a gradual rising tone, \(k = -4\) indicates a rapid falling tone, \(k = -2\) indicates a gradual falling tone, and \(k = 0\) corresponds to a level tone. In Figure \ref{case} (c-g), we modified the reference pitch shape segment of the last word ``moment".

We observe that when a level tone is provided as a reference, the pitch contour corresponding to the word ``moment" in the synthesized speech remains essentially flat. Conversely, when rising or falling tones are used as references, the pitch contour for ``moment" exhibits the corresponding upward or downward movement. Additionally, the pitch contour slopes in Figures \ref{case} (d) and (e) are noticeably steeper than those in Figures \ref{case} (g) and (f). This indicates that the pitch contour in the synthesized speech accurately reflects the same degree of reference rapid or gradual shape. These results demonstrate that our proposed Terminal Intonation Encoder effectively captures the reference intonation pattern, allowing ProsodyFM to achieve precise and fine-grained control over intonation.

\subsubsection{Phrasing Control} For the controllability of the phrase break, we manually add or remove a phrase break and synthesize the corresponding speech using ProsodyFM (we modify the ``Phrase Breaks (last words)" in the Phrase Break Encoder in Figure 3 (b)). In Figure \ref{case} (h), we added a phrase break after the word ``for," and in Figure \ref{case} (i), we removed the phrase break between ``over" and ``stared".

We observe that when a break is added after ``for", as shown in Figure \ref{case} (h), the spectrogram of the synthesized speech displays a noticeable blank space between ``for" and ``a moment." As shown in Figure \ref{case} (i), when the break between ``over" and ``stared" is removed, a previously existing blank space disappears. This demonstrates that ProsodyFM exhibits excellent control over phrasing.

\section{Conclusion}
We proposed ProsodyFM, a novel prosody-aware TTS model designed to enhance phrasing and intonation without requiring any prosodic labels, resulting in more intelligible synthesized speech. We addressed the intonation modeling issue by employing a novel Pitch Processor to highlight pitch shapes and training a bank of intonation shape tokens to model perceptually aligned intonation patterns instead of absolute pitch values. We tackled the break labeling issue by designing a Phrase Break Encoder to capture initial phrase break locations and then adjusting the variable break durations with a Duration Predictor. Our performance experiments demonstrated that ProsodyFM effectively improved the phrasing and intonation aspects of prosody, thereby enhancing overall intelligibility compared to four SOTA models. Our out-of-distribution experiments showed that this enhanced prosody further brought ProsodyFM strong generalizability on unseen complex sentences. Our ablation study verified the effectiveness of our proposed modules. Our case study visually demonstrated ProsodyFM’s powerful, precise and fine-grained control over phrasing and intonation.

\bibliography{aaai25}

\section{Appendix}

\subsection{A. Formulation and Training Algorithm of ProsodyFM}
The training objective of ProsodyFM is to find the network parameters \(\theta\) and the monotonic alignment \(A\) that maximize the log-likelihood of the data samples \(x_1\), as in Equation \ref{formal_loss}. 

\begin{equation}
\max_{\theta, A} L(\theta, A) = \max_{\theta, A} \log p_1(x_1 | c; \theta, A)
\label{formal_loss}
\end{equation}

Due to the computational intractability to find the global solution, we apply the iterative approach introduced in Glow-TTS \cite{kim2020glow}, namely, each iterative of optimization consists of two steps: (\(1\)) searching for the most probable monotonic alignment \( \hat{A} \) given fixed model parameters \(\theta \), and (\(2\)) updating the model parameters \(\theta\) to maximize the log-likelihood. 

For step (\(1\)), following Glow-TTS \cite{kim2020glow}, we implement the MAS algorithm to find the current optimal alignment \(\hat{A}\) between text and speech, as in Equation \ref{prior_A}.

\begin{equation}
\hat{A} = \arg \max_A \sum_{j=1}^{T_{\text{mel}}} \log \mathcal{N}(z_j; \mu_{A(j)}, \sigma_{A(j)})
\label{prior_A}
\end{equation}

For step (\(2\)), following our backbone MatchaTTS \cite{mehta2024matcha}, we train the ProsodyFM with the OT-CFM \cite{lipman2022flow} procedure. So maximizing the log-likelihood of Equation \ref{formal_loss} can be transformed into regressing the ordinary differential equation (ODE) vector field that defines a mapping from a random sample \(x_0\) to a real Mel-spectrogram \(x_1\).

\subsubsection{OT-CFM Loss}

The OT-CFM procedure introduced in \cite{lipman2022flow} operates as an unconditional generative modeling. However, to meet the requirements of ProsodyFM, it needs to be extended to a conditional generative modeling. ProsodyFM leverages target text alongside the speaker and prosody information as condition \(c\) to guide the speech synthesis process. The objective of this conditional generative modeling is to produce new Mel-spectrogram samples \(x_1\) that are approximately distributed according to \(p_1(x_1|c)\), where both \(x_1\) and \(c\) are random variables. This task can be reformulated as training a neural network to model \( p_1(x_1 | c_1) \), where \( c_1 \sim p(c) \). By providing the neural network with sufficient samples \( c_1 \) from the dataset, and using parameter sharing across conditions, the network effectively approximates \( p_1(x_1 | c) \). 

Given a particular Mel-spectrogram sample \(x_1 \sim p_1\), it is associated with a specific condition \(c_1 \sim p(c)\). We can replace the random variable \(x_0\), \(x_t\), and \(x_1\)  in the unconditional OT-CFM to \(x_0|c_1\), \(x_t|c_1\), and \(x_1|c_1\), where \(x_0, x_t, x_1\) are random variables and \(c_1\) is the specific condition sample. Then we can construct the Gaussian conditional probability path as:

\begin{equation}
\begin{split}
&p_{t|1}(x_t \mid x_0, x_1, c_1) = \\
&\mathcal{N}(x_t; \left(1 - (1 -\sigma_{\text{min}})t\right)x_0 + t x_1, \sigma_{min}^2I ) 
\end{split}
\label{conditional_probability_path}
\end{equation}

The objective of the conditional OT-CFM for ProsodyFM is: 

\begin{equation}
\begin{split}
\mathcal{L}_{\text{OT-CFM}} =& \mathbb{E}_{t \sim \mathcal{U}[0,1], x_1 \sim p_1, x_0 \sim p_0, x_t \sim p_{t|1}(x_t \mid x_0, x_1, c_1)} \\
&\left[ \left\lVert u_\theta(x_t, c_1, t) - [x_1-(1-\sigma_{min})x_0] \right\rVert^2 \right]
\end{split}
\label{OT_CFM_loss}
\end{equation}

In this particular sample \((x_1, c_1)\), the Flow Prediction Decoder (vector field estimator) \(u_\theta\) is trained to estimate the conditional vector field \(u_t(x_t|c_1)\). By providing it with sufficiently diverse data samples \(x_1\) with \(c_1\) from the dataset and parameter sharing, the \(u_\theta\) approximates \( u_t(x_1 | c) \) which is the real vector field that generates the target probability path \(p_t(x_t|c)\). By solving the joint ODE of Equation \ref{joint_ode} during inference, we can obtain the Mel-spectrogram \(x_1\) given any \(c\) as a condition.

\begin{equation}
\frac{d}{dt} 
\begin{pmatrix}
x_t \\
\log p_t(x_t)
\end{pmatrix}
=
\begin{pmatrix}
u_\theta(x_t) \\
-\left( \nabla \cdot u_\theta \right)(x_t)
\end{pmatrix}
\label{joint_ode}
\end{equation}

\subsubsection{Prior Loss}

We introduce a prior loss \(\mathcal{L}_{prior}\) as an explicit maximum log-likelihood objective for the conditional encoder. The conditional encoder, denoted as \(f_{c\_enc}\), include the Text, Phrase Break, Terminal Intonation, Speaker, and Fusion encoders. Intuitively, providing a condition that is closely related to the generation target as input can reduce the training burden on the vector field estimator. The prior loss  \(\mathcal{L}_{prior}\) is given by:

\begin{equation}
\begin{split}
\mathcal{L}_{prior} &= -\sum_{j=1}^{T_{mel}} \log\mathcal{N}(z_j; \mu_{{\hat{A}}(j)}, I) \\ &= -\sum_{j=1}^{T_{mel}} (\frac{n}{2} \log (2\pi) + \frac{1}{2} \|z_j - \mu_{{\hat{A}}(j)}\|^2)
\end{split}
\label{prior_loss}
\end{equation}

where \(z = f_{c\_enc}(w, x_1)\) is the output of \(f_{c\_enc}\), \(z_j\) is the \(j\)-th frame of the \(z\), and \(n\) represents the dimensionality of \(z_j\). The difference between Equation \ref{prior_A} and Equation \ref{prior_loss} is that Equation \ref{prior_A} aims to find the optimal alignment \(\hat{A}\) given the current \(f_{c\_enc}\) parameters, while Equation \ref{prior_loss} seeks to optimize the \(f_{c\_enc}\) parameters with \(\hat{A}\) fixed.

\subsubsection{Duration Loss}

During training, the transcript of reference speech is matched with the target text. So we can directly perform the MAS algorithm introduced in Glow-TTS \cite{kim2020glow} to search for the most probable alignment \(A^*\) between phone-level prior statistics and the frame-level speech representation. However, during inference, the input reference speech may not match the target text. To estimate the best monotonic alignment \(A^*\) at inference, we need to design a During Predictor. 

As shown in Figure \ref{model}, we incorporate the Duration Predictor \(f_{dur}\) on top of the Fusion Encoder \(f_{fus}\). It follows the architecture of the Duration Predictor in MatchaTTS. It is trained with the mean squared error loss (MSE) in the logarithmic domain, as described in Equation \ref{loss_dur}. To avoid the gradient affecting the maximum log-likelihood objective (Equation \ref{formal_loss}), we stop the gradient propagation between \(f_{fus}\) and \(f_{dur}\). 

\begin{equation}
\begin{split}
\mathcal{L}_{dur} &= MSE(f_{dur}, [d_1, ...,d_i, ..., d_{T_{text}}]), \\  
d_i &= \sum_{j=1}^{T_{\text{mel}}} 1_{A^*(j) = i}, \quad i=1, \dots, T_{\text{text}}
\end{split}
\label{loss_dur}
\end{equation}

\subsubsection{Text-pitch Alignment Loss}
In Text-Pitch Aligner, we aim to align the BERT-derived word embedding of the last word \(e_k\) with the corresponding reference intonation feature \(r_k\) in the same feature space. We use the L2 loss as the alignment loss \(\mathcal{L}_{tp\_align}\) (Equation \ref{tp_loss}). The reference encoder is detached from the computation graph when computing this loss.

\begin{equation}
\mathcal{L}_{tp\_align} = -\sum_{k=1}^{T_{lword}} \|\mathbf{e}_k - \mathbf{r}_k\|_2^2,
\label{tp_loss}
\end{equation}
where \(T_{lword}\) is the number of last words.

\subsubsection{Training Algorithm}
\begin{algorithm}[t]
\caption{Training Algorithm of ProsodyFM}
\begin{algorithmic}[1] 
 \Require the target text $w$, the reference Mel-spectrogram $x_1$, the Mel-spectrogram length $T_{mel}$, the text length $T_{text}$
 \Ensure optimal vector field predictor $u_{\theta^*}$

\Function{TrainStep}{$w$, $x_1$, $x_0$}
    \State Input the target text $w$ and the reference Mel-spectrogram $x_1$ into the conditional encoder \(f_{c\_enc}\) and obtain the output prior statistics $\mu_i$;
    
    \State Search for the optimal monotonic alignment $\hat{A}$ using MAS (Equation \ref{prior_A});
    
    \State Align the prior statistics $\mu_i$ with the reference Mel-spectrogram $x_1$ based on $\hat{A}$ to get the condition $c_1$;

    \State Compute \(\mathcal{L}_{prior}\), \(\mathcal{L}_{dur}\) and \(\mathcal{L}_{tp\_align}\) according to Equation \ref{prior_loss}, Equation \ref{loss_dur} and Equation \ref{tp_loss};

    \State Sample time $t \sim \text{Uniform}[0, 1]$;
    
    \State Sample $x_t \sim \mathcal{N}(x_t; \left(1 - (1 -\sigma_{\text{min}})t\right)x_0 + t x_1, \sigma_{min}^2I )$;
    
    \State Compute \(\mathcal{L}_{OT-CFM}\) according to Equation \ref{OT_CFM_loss};

    \State Gradient descent on the total loss \(\mathcal{L}_{ProsodyFM}\) (Equation \ref{total_loss}) and obtain new model parameters \(\hat{\theta}\)
\EndFunction

\While{Perform Training}
    \State Take a batch of target text $w$ and the reference Mel-spectrogram $x_1$;
    \State Sample $x_0 \sim \mathcal{N}(0, I)$;
    \State Call \Call{TrainStep}{$w$, $x_1$, $x_0$};
\EndWhile

\end{algorithmic}
\label{prosodyfm_training_algorithm}
\end{algorithm}

The four losses are marked with double lines in Figure \ref{model} and Figure \ref{model_components}. The total loss for training ProsodyFM is:
\begin{equation}
\mathcal{L}_{ProsodyFM} = \mathcal{L}_{OT-CFM} + \mathcal{L}_{prior} +
\mathcal{L}_{dur} +
\mathcal{L}_{tp\_align}
\label{total_loss}
\end{equation}
We present the training algorithm of ProsodyFM in Algorithm \ref{prosodyfm_training_algorithm}.

\subsection{B. Performance of the Phrase Break Predictor}

\begin{table}[t]
\resizebox{\columnwidth}{!}{%
\begin{tabular}{c|ccc}
\toprule[1.5pt]
\multicolumn{1}{l|}{}         & Precision & Recall  & F1      \\
\midrule[1pt]
LibriTTS (w final-break)    & 93.96\%   & 86.36\% & 90.00\% \\
LibriTTS (w\_o final-break) & 91.04\%   & 80.51\% & 85.45\% \\
\midrule[1pt]
VCTK (w final-break)        & 99.15\%   & 92.86\% & 95.90\% \\
VCTK (w\_o final-break)     & 94.74\%   & 66.67\% & 78.26\% \\
\bottomrule[1.5pt]
\end{tabular}%
}
\caption{The performance of the phrase break predictor on the LibriTTS and VCTK validation sets. ``w final-break" and ``w\_o final-break" refer to the inclusion or exclusion of the sentence-final break, respectively, when calculating precision, recall, and F1 scores.}
\label{phrase_break_predictor_performance}
\end{table}

The Phrase Break Predictor is trained independently from ProsodyFM on the LibriTTS training set. We fine-tune T5 and consider the phrase breaks obtained from the PSST as the ground truth labels when fine-tuning. We present its performance on the LibriTTS and VCTK validation sets in Table ~\ref{phrase_break_predictor_performance}. Given that each sentence inherently concludes with a sentence-final break, we also report results excluding sentence-final breaks. We can observe a noticeable performance decline on the VCTK dataset when these final breaks are excluded. This decline can be attributed to the relatively short sentences in VCTK, which contain fewer intra-sentence breaks.

\subsection{C. More Examples for the Prosody Controllability}

\begin{figure*}[t]
\centering
\begin{minipage}{0.5\textwidth}
    \centering
    \includegraphics[width=\linewidth]{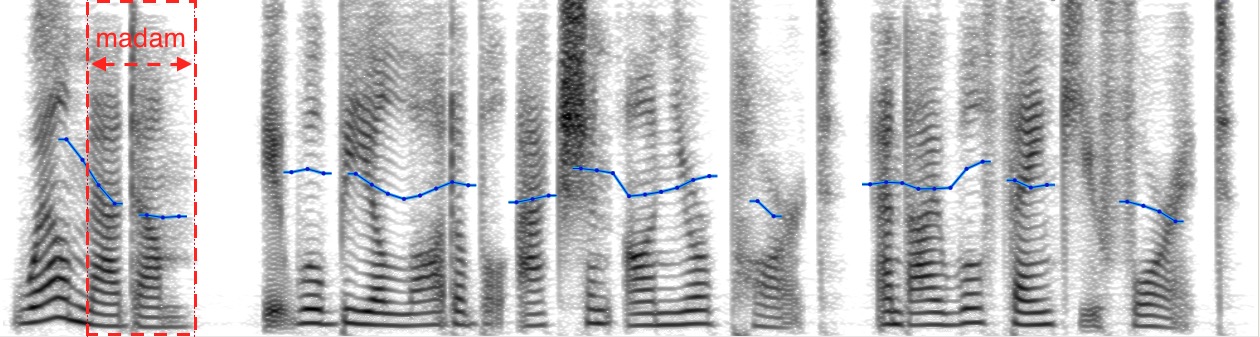}
    \caption*{(a) ProsodyFM}
\end{minipage}\hfill
\begin{minipage}{0.5\textwidth}
    \centering
    \includegraphics[width=\linewidth]{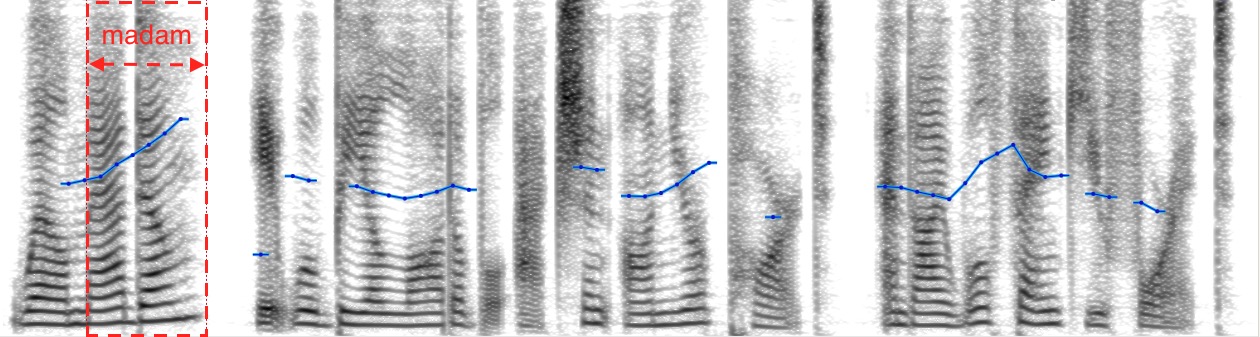}
    \caption*{(b) Rising k=+2}
\end{minipage}\hfill
\begin{minipage}{0.5\textwidth}
    \centering
    \includegraphics[width=\linewidth]{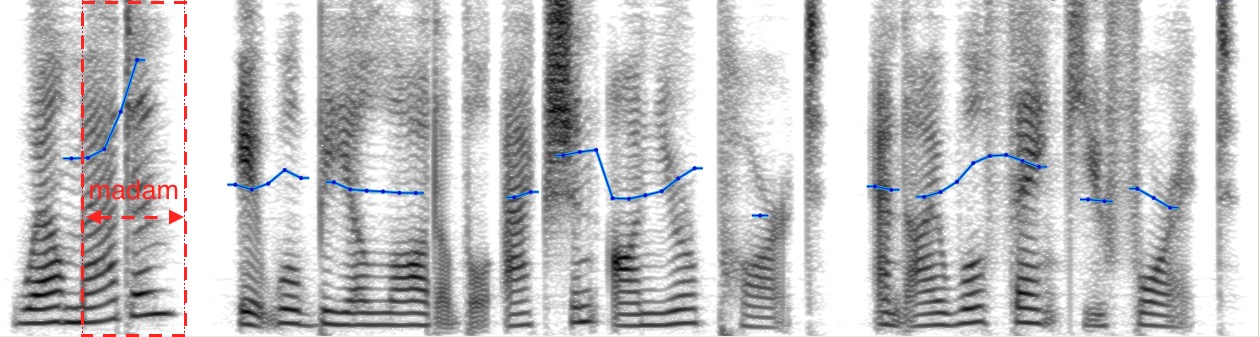}
    \caption*{(c) Rising k=+4}
\end{minipage}\hfill
\begin{minipage}{0.5\textwidth}
    \centering
    \includegraphics[width=\linewidth]{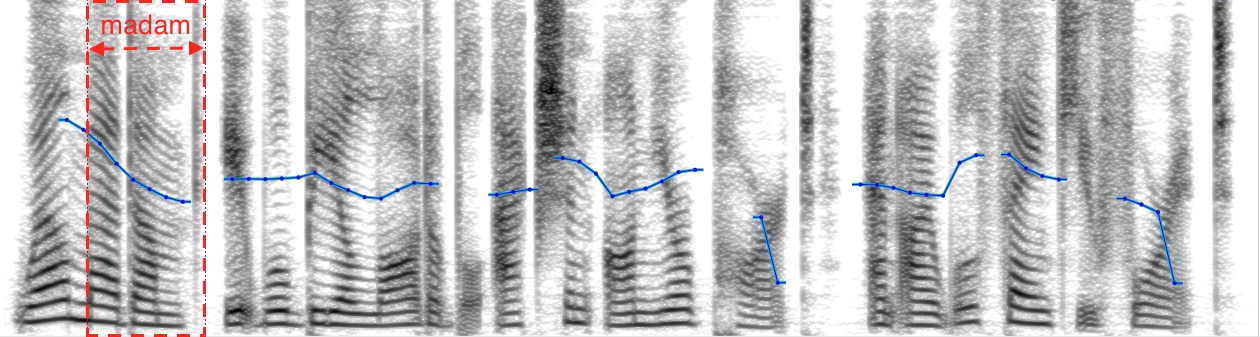}
    \caption*{(d) Falling k=-2}
\end{minipage}\hfill
\begin{minipage}{0.5\textwidth}
    \centering
    \includegraphics[width=\linewidth]{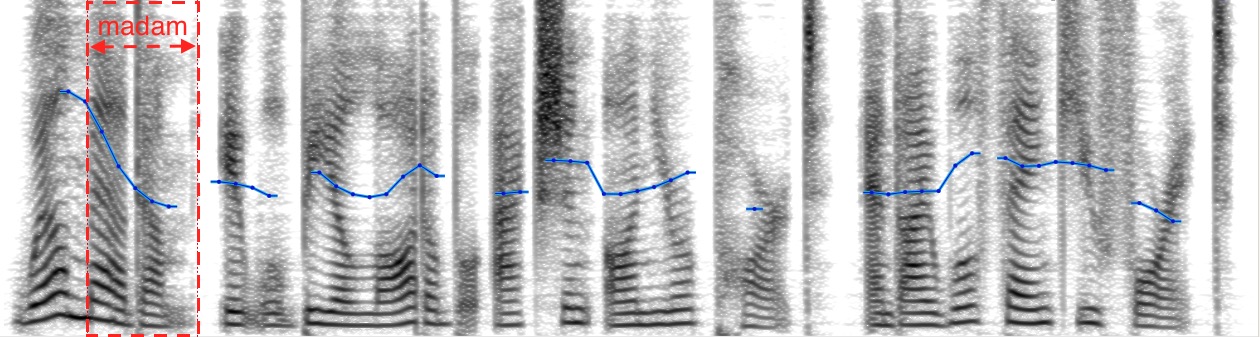}
    \caption*{(e) Falling k=-4}
\end{minipage}\hfill
\begin{minipage}{0.5\textwidth}
    \centering
    \includegraphics[width=\linewidth]{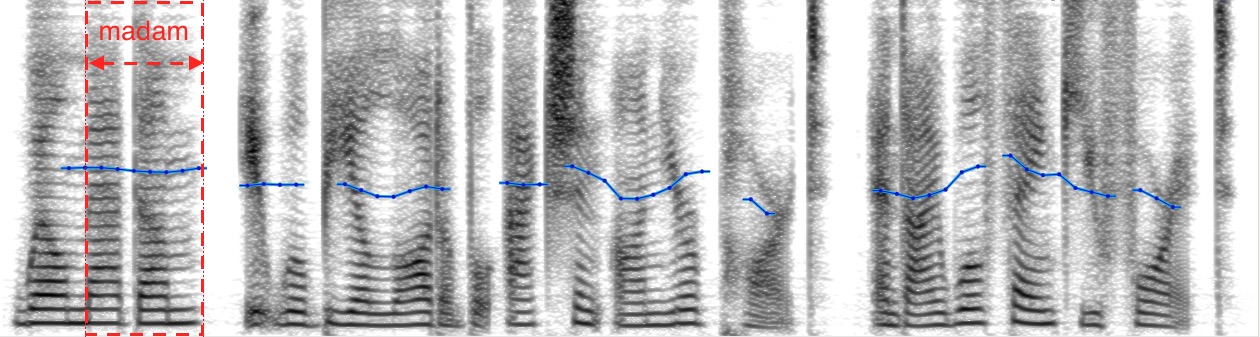}
    \caption*{(f) Level k=0}
\end{minipage}\hfill
\caption{The terminal intonation control results for the 7302\_86815\_000052\_000000.wav.}
\label{into}
\end{figure*}

\begin{figure*}[t]
\centering
\begin{minipage}{0.5\textwidth}
    \centering
    \includegraphics[width=\linewidth]{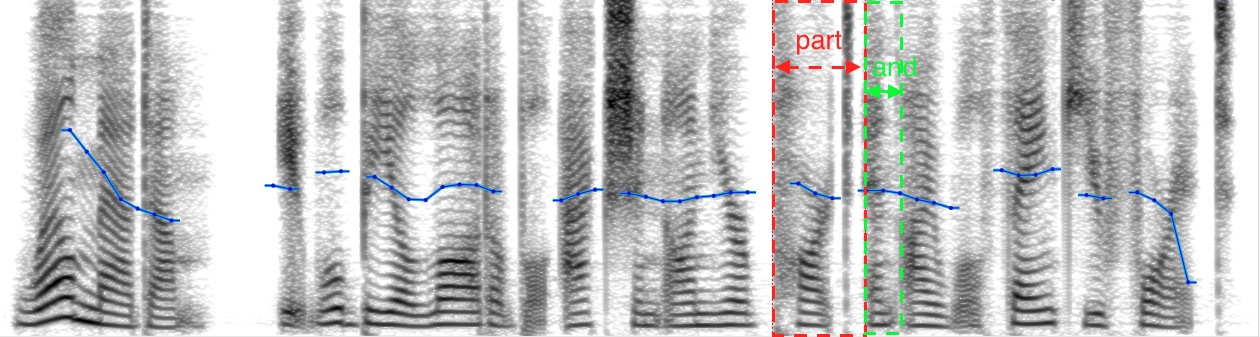}
    \caption*{(g) Remove a break}
\end{minipage}\hfill
\begin{minipage}{0.5\textwidth}
    \centering
    \includegraphics[width=\linewidth]{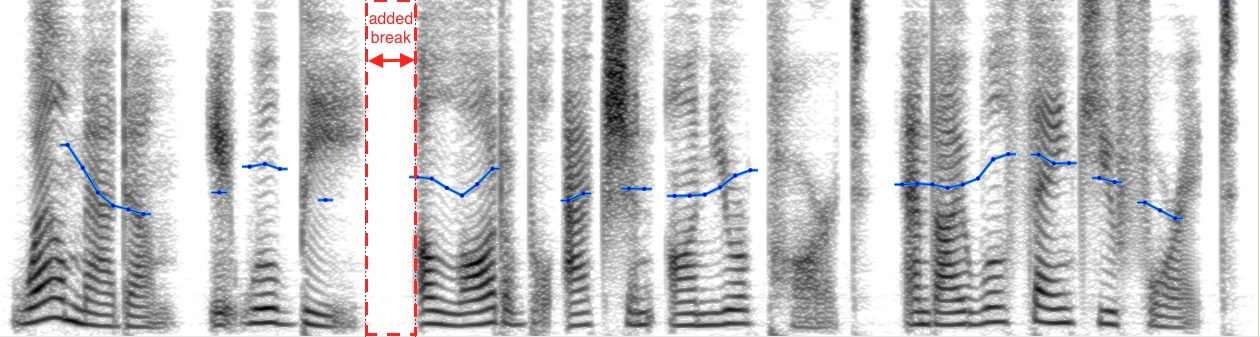}
    \caption*{(h) Add a break}
\end{minipage}\hfill
\caption{The phrase break control results for the 7302\_86815\_000052\_000000.wav.}
\label{break}
\end{figure*}

\begin{figure*}[t]
\centering

\begin{minipage}{0.5\textwidth}
    \centering
    \includegraphics[width=\linewidth]{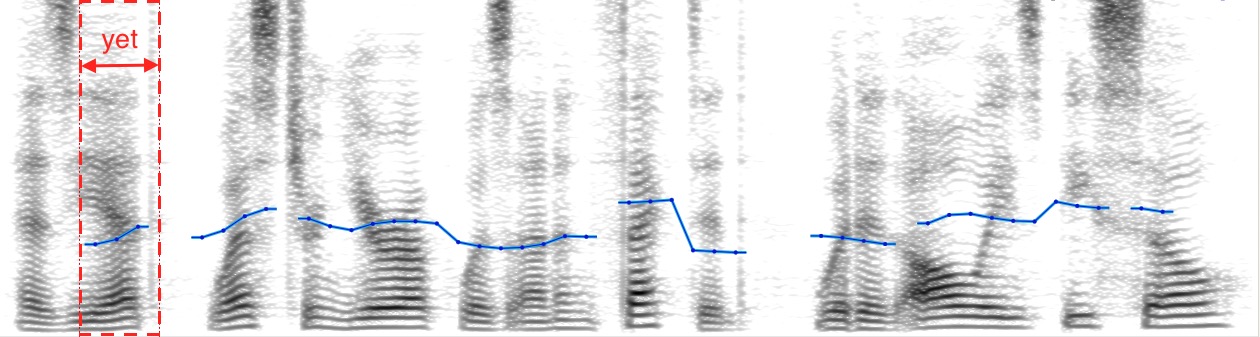}
    \caption*{(a) ProsodyFM}
\end{minipage}\hfill
\begin{minipage}{0.5\textwidth}
    \centering
    \includegraphics[width=\linewidth]{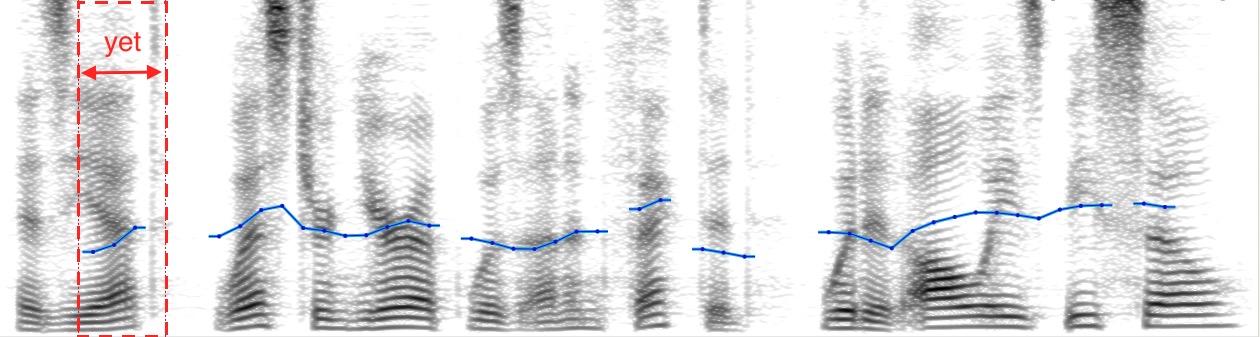}
    \caption*{(b) Rising k=+2}
\end{minipage}\hfill
\begin{minipage}{0.5\textwidth}
    \centering
    \includegraphics[width=\linewidth]{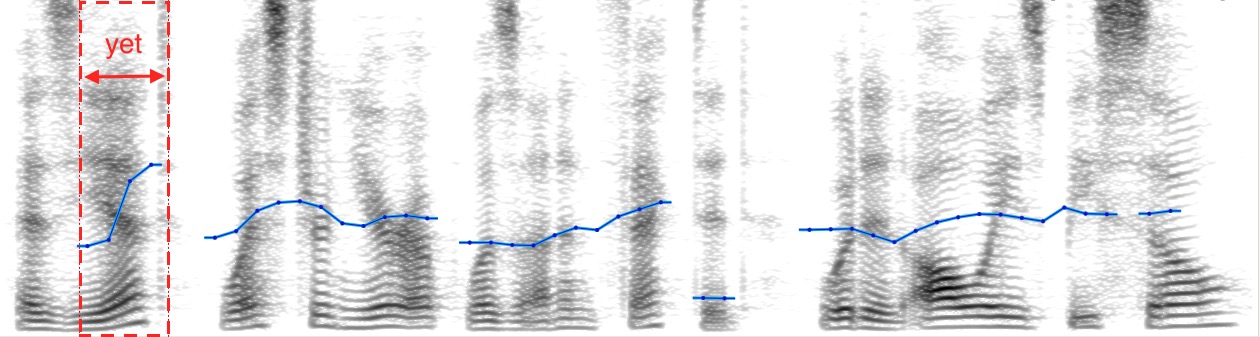}
    \caption*{(c) Rising k=+4}
\end{minipage}\hfill
\begin{minipage}{0.5\textwidth}
    \centering
    \includegraphics[width=\linewidth]{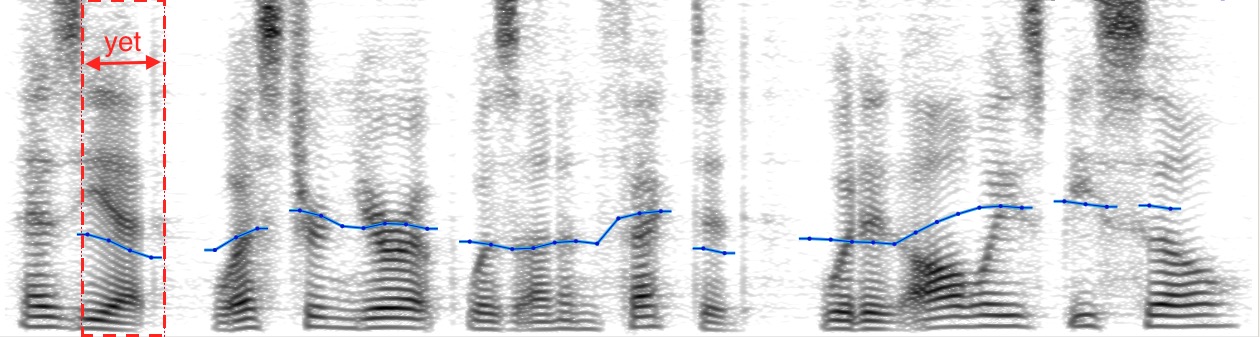}
    \caption*{(d) Falling k=-2}
\end{minipage}\hfill
\begin{minipage}{0.5\textwidth}
    \centering
    \includegraphics[width=\linewidth]{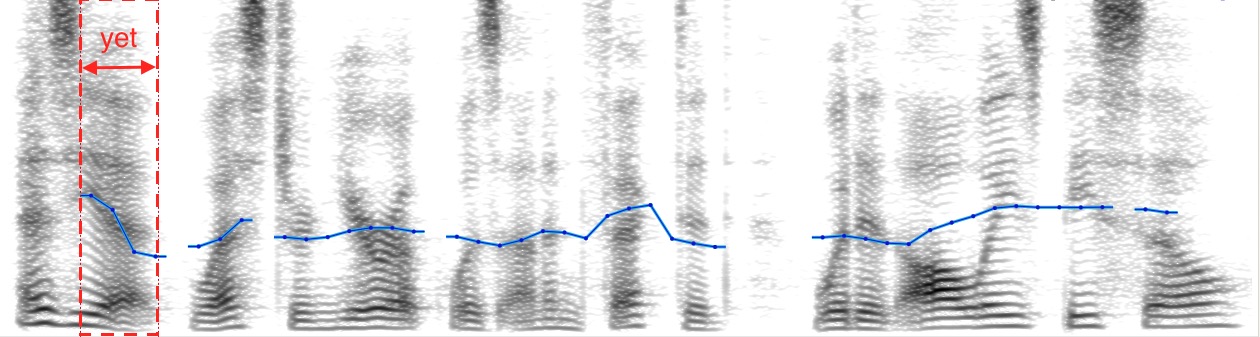}
    \caption*{(e) Falling k=-4}
\end{minipage}\hfill
\begin{minipage}{0.5\textwidth}
    \centering
    \includegraphics[width=\linewidth]{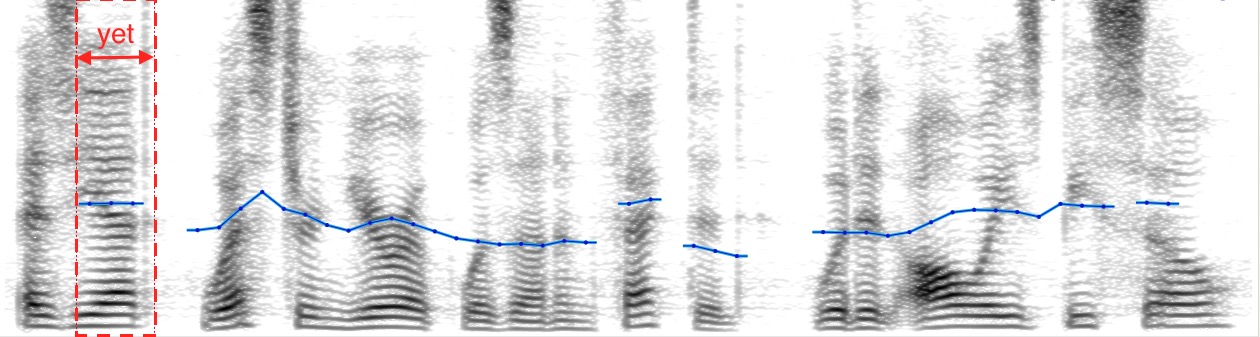}
    \caption*{(f) Level k=0}
\end{minipage}\hfill
\caption{The terminal intonation control results for the 2002\_139469\_000016\_000005.wav.}
\label{into_1}
\end{figure*}

\begin{figure*}[t]
\centering
\begin{minipage}{0.5\textwidth}
    \centering
    \includegraphics[width=\linewidth]{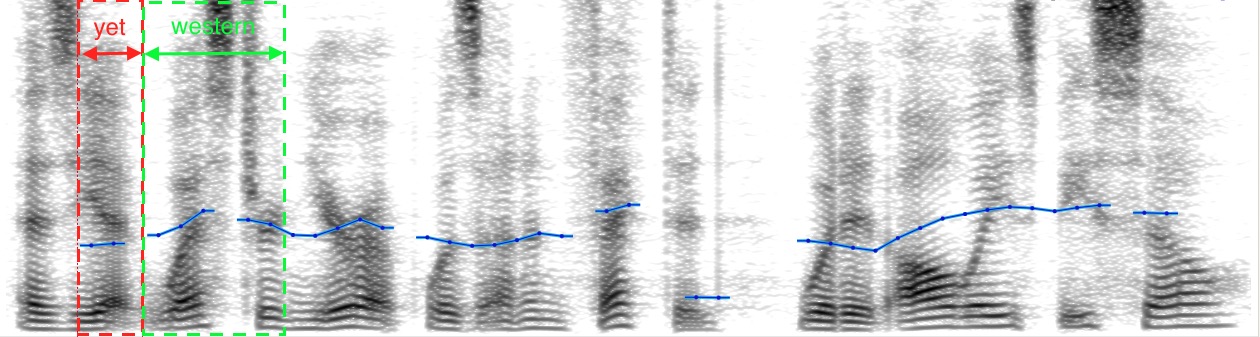}
    \caption*{(g) Remove a break}
\end{minipage}\hfill
\begin{minipage}{0.5\textwidth}
    \centering
    \includegraphics[width=\linewidth]{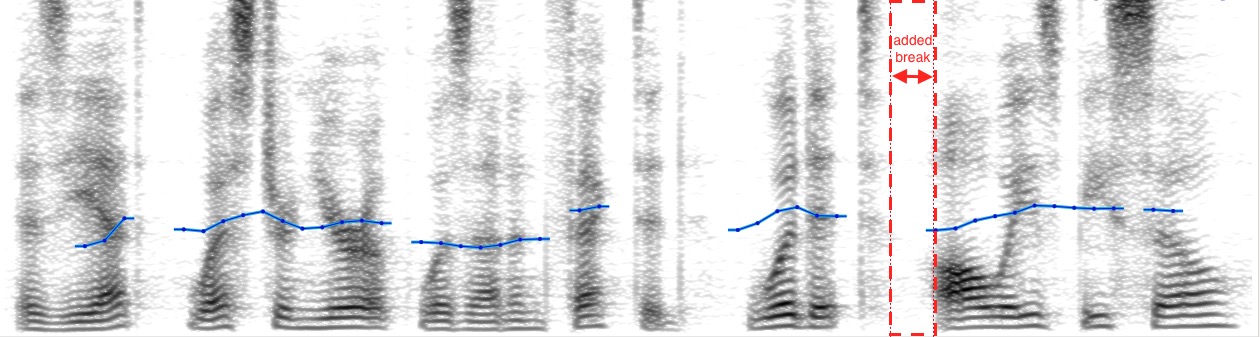}
    \caption*{(h) Add a break}
\end{minipage}\hfill
\caption{The phrase break control results for the 2002\_139469\_000016\_000005.wav.}
\label{break_1}
\end{figure*}

Figure \ref{into} and Figure \ref{break} show the intonation and phrasing control results of a speech sample 7302\_86815\_000052\_000000.wav from the LibriTTS testing set. The original labeled transcript is ``Well madam (falling tone) it will be a laudable action on your part (falling tone) and I will thank you for it (falling tone)". For the intonation control, in Figure \ref{into} (b-f), we linearly adjust the reference pitch shape segment of the last word ``madam" with different slope. For the phrasing control, in Figure \ref{break} (g) we remove the break between ``part" and ``and". In Figure \ref{break} (h) we add a break after ``be". 

Figure \ref{into_1} and Figure \ref{break_1} show the intonation and phrasing control results of a speech sample 2002\_139469\_000016\_000005.wav from the LibriTTS testing set. The original labeled transcript is ``As yet (rising tone) western Europe was uninfected (falling tone) would it always be so (rising tone)". For the intonation control, in Figure \ref{into_1} (b-f), we linearly adjust the reference pitch shape segment of the last word ``yet" with different slope. For the phrasing control, in Figure \ref{break_1} (g) we remove the break between ``yet" and ``western". For Figure \ref{break_1} (h), we add a break after ``it". 

These two samples and more speech examples for the prosody controllability experiment can be found on our demo page. We strongly recommend you listen to the speech samples.

\subsection{D. Instructions for the MOS Test}
\begin{figure*}[t]
  \centering
\begin{minipage}{0.99\textwidth}
    \centering
    \includegraphics[width=0.65\linewidth]{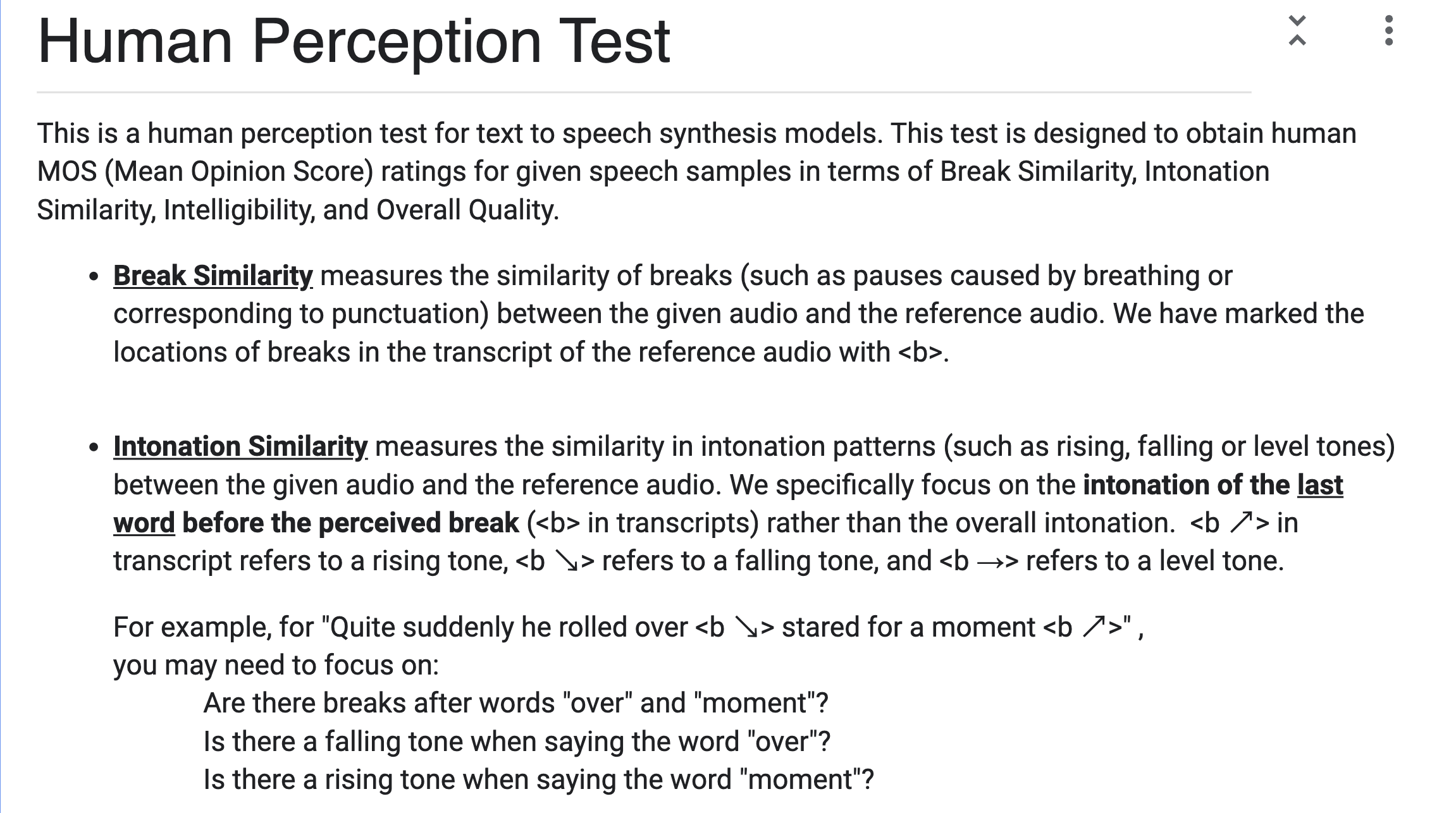}
\end{minipage}\vfill
\begin{minipage}{0.99\textwidth}
    \centering
    \includegraphics[width=0.65\linewidth]{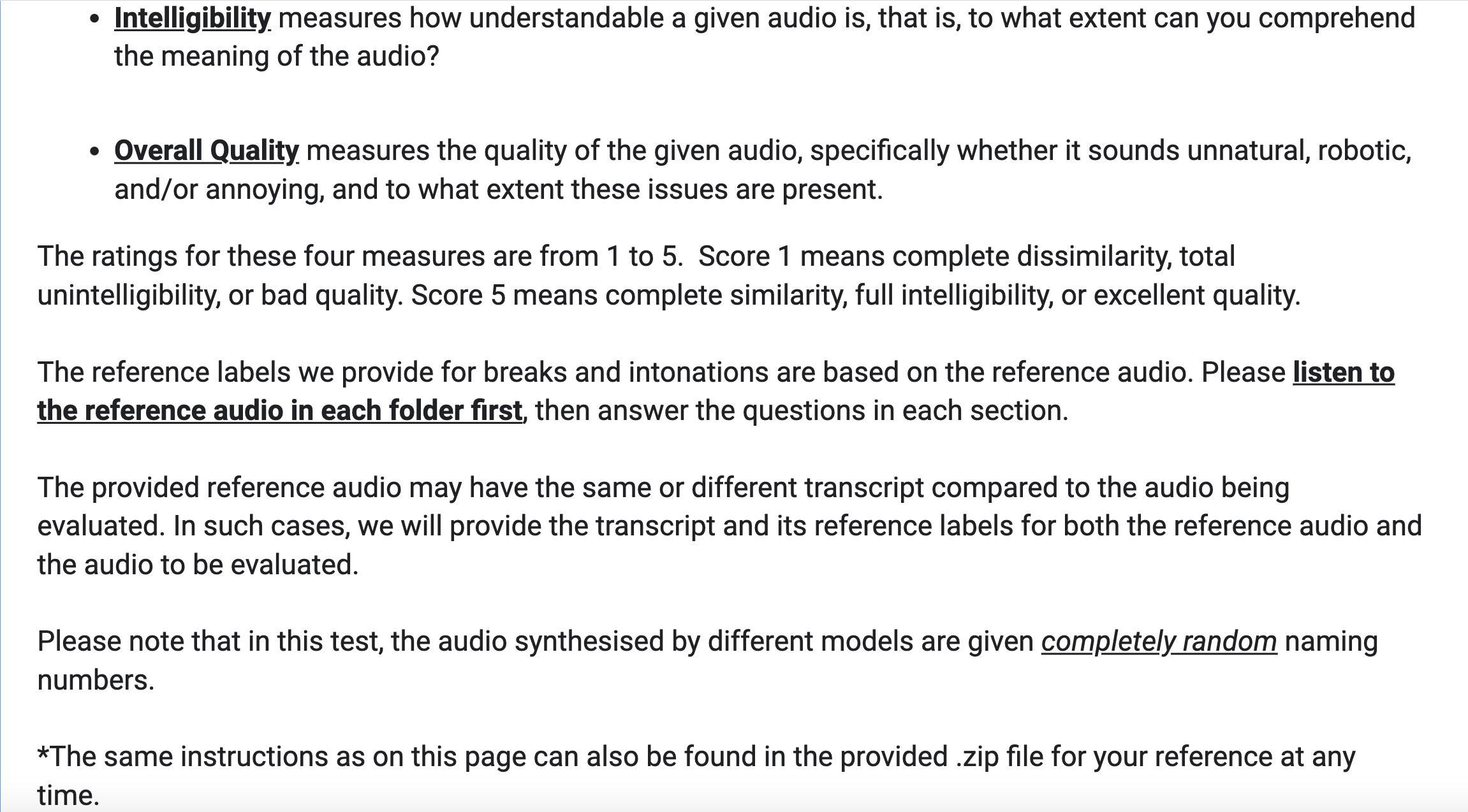}
\end{minipage}
\caption {The instruction page of our Mean Opinion Score human listening test.}
  \label{mos_instruction}
\end{figure*}
Figure~\ref{mos_instruction} shows the instruction page of our crowd-sourced Mean Opinion Score human listening test. These instructions are given to the testers prior to the commencement of their rating task.

\subsection{E. Labeled Transcripts of the MOS Test}
\begin{figure*}[t!]
  \centering
\begin{minipage}{0.99\textwidth}
    \centering
    \includegraphics[width=0.8\linewidth]{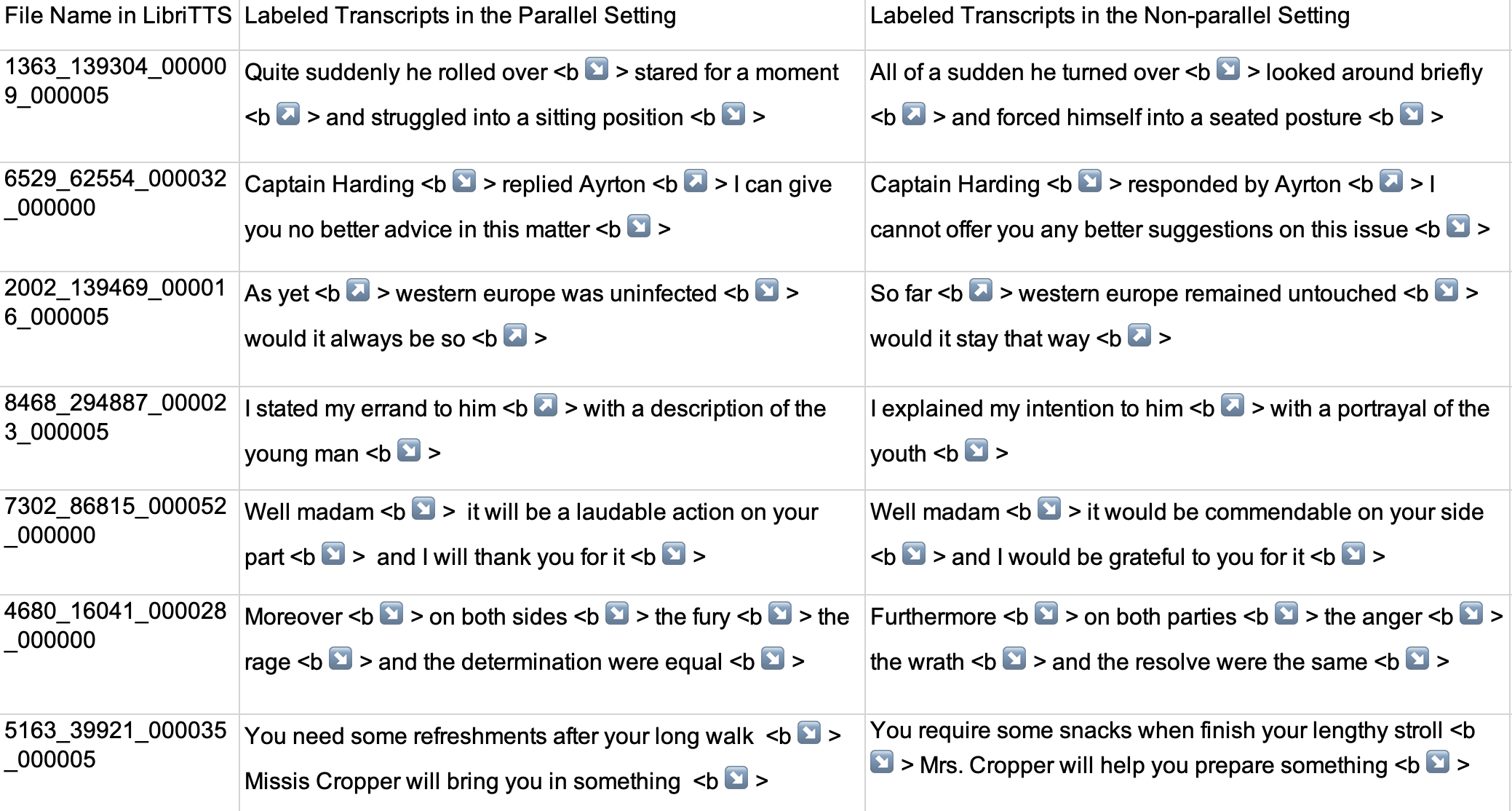}
\end{minipage}\vfill
\begin{minipage}{0.99\textwidth}
    \centering
    \includegraphics[width=0.8\linewidth]{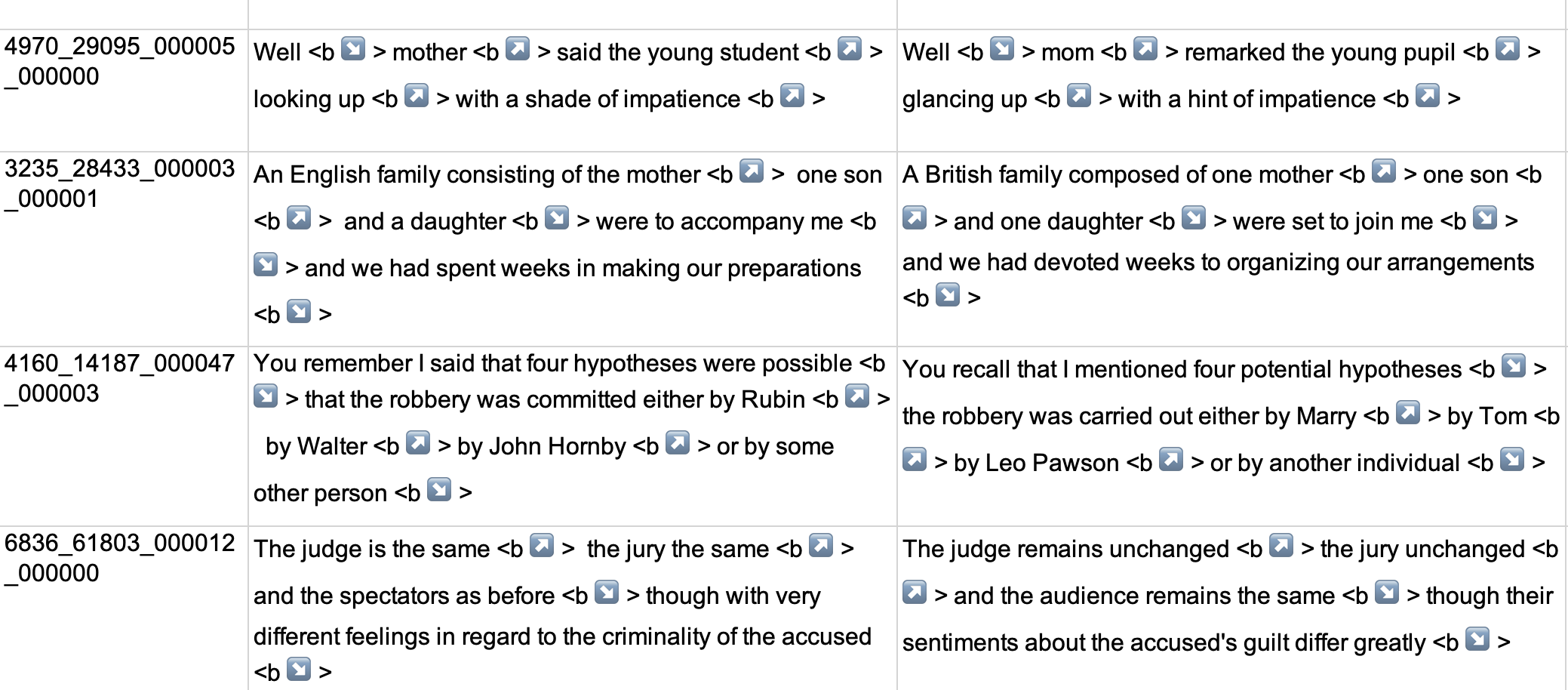}
\end{minipage}\vfill
\begin{minipage}{0.99\textwidth}
    \centering
    \includegraphics[width=0.8\linewidth]{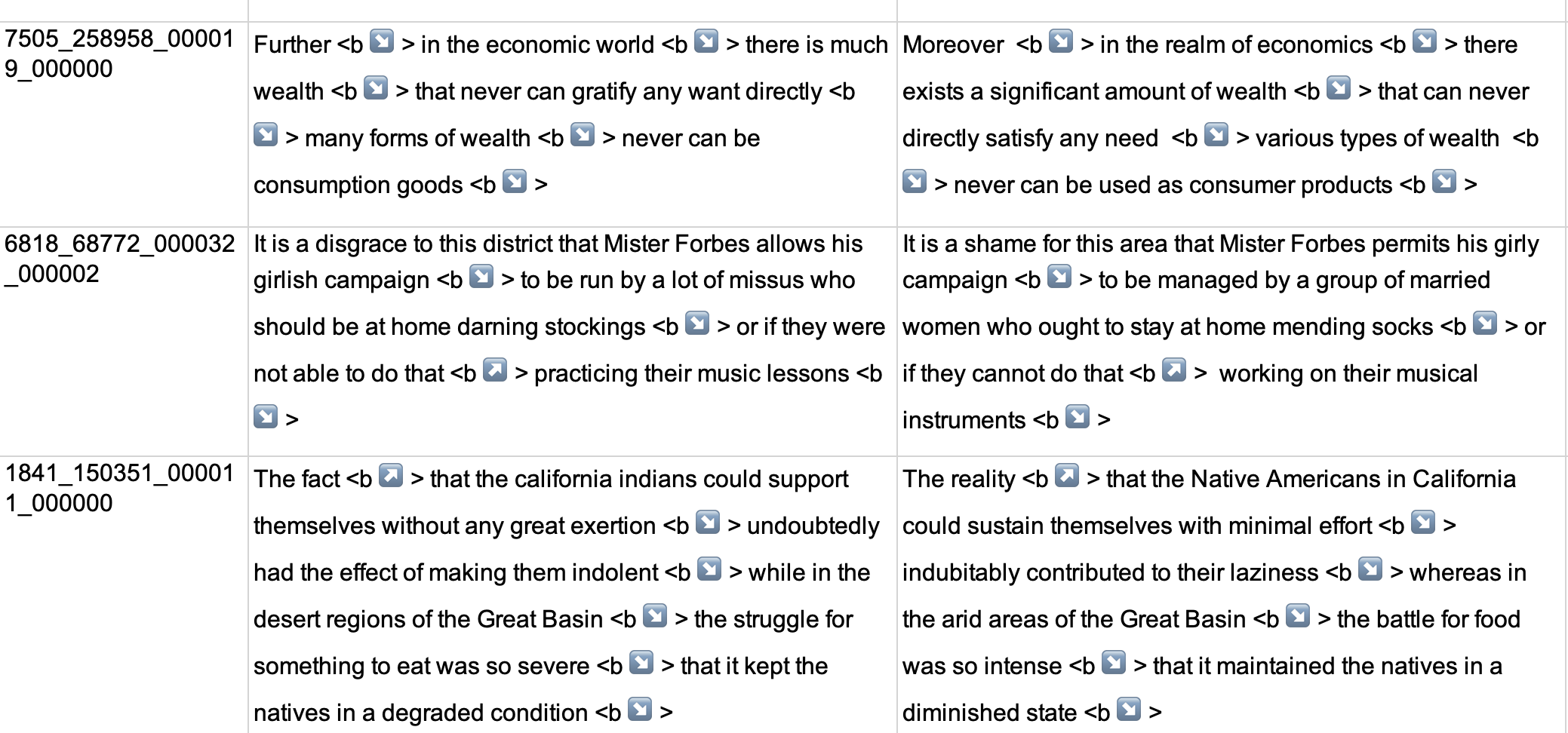}
\end{minipage}\vfill
\begin{minipage}{0.99\textwidth}
    \centering
    \includegraphics[width=0.8\linewidth]{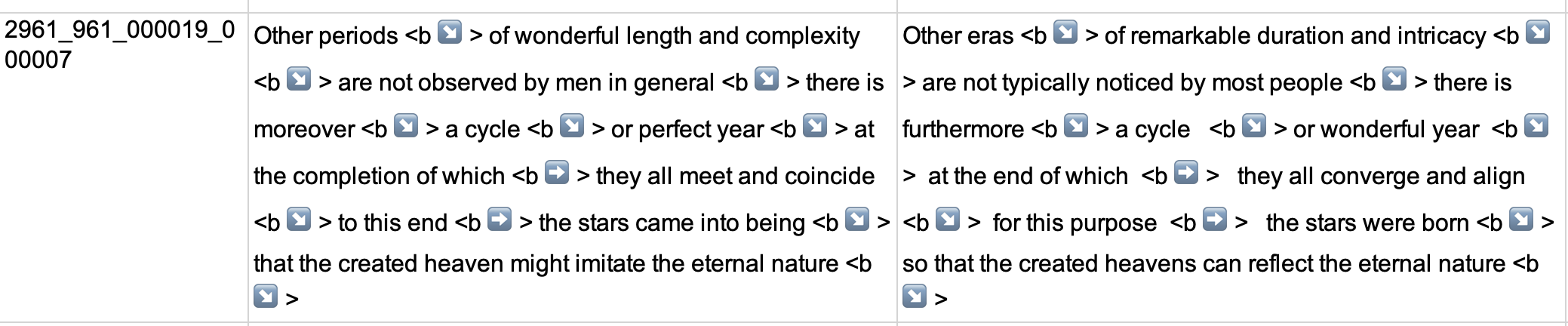}
\end{minipage}
\caption {The labeled transcripts provided in the human listening test for all 15 testing samples under parallel and non-parallel settings.}
\label{mos_transcripts}
\end{figure*}

Figure~\ref{mos_transcripts} presents the labeled transcripts provided in the human listening test for 15 random samples under parallel and non-parallel settings. The $\textless$ b $\nearrow$ $\textgreater$, $\textless$ b $\searrow$ $\textgreater$, and $\textless$ b $\rightarrow$ $\textgreater$ describe a phrase break with rising, falling, and level intonation on the last word, respectively.  The terminal intonation and phrase break labels we provided are based on the actual pitch contour of the speech signal, supplemented by reference to the perceptual judgments of two human annotators.

\end{document}